\documentclass[10pt,journal,compsoc]{IEEEtran}
\usepackage{amsmath}
\usepackage[utf8]{inputenc} % allow utf-8 input
\usepackage[T1]{fontenc}    % use 8-bit T1 fonts
\usepackage{hyperref}       % hyperlinks
\usepackage{url}            % simple URL typesetting
\usepackage{booktabs}       % professional-quality tables
\usepackage{amsfonts}       % blackboard math symbols
\usepackage{microtype}      % microtypography
\usepackage{xcolor}         % colors

\DeclareMathAlphabet\mathbfcal{OMS}{cmsy}{b}{n}
\usepackage{algorithm, algorithmic}
\usepackage{amssymb}
\usepackage{graphicx}
\usepackage{textcomp}
\usepackage{bbold}
\usepackage{balance}
\usepackage{bm}
\usepackage{cuted}

\ifCLASSOPTIONcompsoc
  \usepackage[nocompress]{cite}
\else
  \usepackage{cite}
\fi

\usepackage{xspace}
\usepackage{soul}
\DeclareMathAlphabet\mathbfcal{OMS}{cmsy}{b}{n}
\makeatletter
\DeclareRobustCommand\onedot{\futurelet\@let@token\@onedot}
\def\@onedot{\ifx\@let@token.\else.\null\fi\xspace}

\def\eg{\emph{e.g}\onedot} 
\def\ie{\emph{i.e}\onedot}

\newcommand{\printfnsymbol}[1]{%
  \textsuperscript{\@fnsymbol{#1}}%
}
\makeatother

\ifCLASSINFOpdf
\else
\fi
\begin{document}
\title{FlowX: Towards Explainable Graph Neural Networks via Message Flows}

\author{Shurui Gui, 
        Hao Yuan,
        Jie Wang,
        Qicheng Lao,
        Kang Li,
        and Shuiwang Ji,~\IEEEmembership{Fellow,~IEEE}% <-this % stops a space
\IEEEcompsocitemizethanks{\IEEEcompsocthanksitem S. Gui, H. Yuan, S. Ji are with the Department of Computer Science and Engineering, Texas A\&M University, College Station, TX 77843, USA, e-mail: \{shurui.gui, hao.yuan, sji\}@tamu.edu.
\IEEEcompsocthanksitem J. Wang is with the Department of Electronic Engineering and Information Science, University of Science and Technology of China, Hefei, Anhui, 230026, China, email: jiewangx@ustc.edu.cn.
\IEEEcompsocthanksitem Q. Lao is with the School of Artificial Intelligence, Beijing University of Posts and Telecommunications, Beijing, 100876, China, email: qicheng.lao@gmail.com.   
\IEEEcompsocthanksitem K. Li is with the West China Biomedical Big Data Center, West China Hospital, Chengdu, Sichuan, 610041, China, email: likang@wchscu.cn}% <-this % stops an unwanted space
}


\markboth{Preprint}%
{Shell \MakeLowercase{\textit{et al.}}: Bare Demo of IEEEtran.cls for Computer Society Journals}



\IEEEtitleabstractindextext{%
\begin{abstract}
We investigate the explainability of graph neural networks (GNNs) as a step toward elucidating their working mechanisms. While most current methods focus on explaining graph nodes, edges, or features, we argue that, as the inherent functional mechanism of GNNs, message flows are more natural for performing explainability. To this end, we propose a novel method here, known as FlowX, to explain GNNs by identifying important message flows. To quantify the importance of flows, we propose to follow the philosophy of Shapley values from cooperative game theory. To tackle the complexity of computing all coalitions' marginal contributions, we propose a flow sampling scheme to compute Shapley value approximations as initial assessments of further training. We then propose an information-controlled learning algorithm to train flow scores toward diverse explanation targets: necessary or sufficient explanations. Experimental studies on both synthetic and real-world datasets demonstrate that our proposed FlowX and its variants lead to improved explainability of GNNs.
\end{abstract}

}

\maketitle

\IEEEdisplaynontitleabstractindextext

% \fi
\IEEEpeerreviewmaketitle

\IEEEraisesectionheading{\section{Introduction}\label{sec:introduction}}
\IEEEPARstart{W}{ith} the advances of deep learning, graph neural networks (GNNs) are achieving promising performance on many graph tasks, including graph classification~\cite{gin,Gao:ICML19, chen2020simple}, node classification~\cite{gcn, gat, wu2019simplifying}, and graph generation~\cite{luo2021graphdf, you2018graph}. Many research efforts have been made to develop advanced graph operations, such as graph message passing~\cite{gcn, gat, li2019deepgcns}, graph pooling~\cite{Yuan:ICLR2020,zhang2018end, ying2018hierarchical}, and 3D graph operations~\cite{schutt2018schnet,liu2021spherical}. Deep graph models usually consist of many layers of these operations stacked on top of each other and interspersed with nonlinear functions. The resulting deep models are usually deep, highly nonlinear, and complex. While these complex systems allow for accurate modeling, their decision mechanisms are highly elusive and not human-intelligible. Given the increasing importance and demand for trustworthy and fair artificial intelligence, it is imperative to develop methods to open the black box and explain these highly complex deep models.
Driven by these needs, significant efforts have been made to investigate the explainability of deep models on images and texts. These methods are developed from different perspectives, including studying the gradients of models~\cite{simonyan2013deep,smilkov2017smoothgrad, yang2019xfake}, mapping hidden features to input space~\cite{zhou2016learning, selvaraju2017grad}, occluding different input features~\cite{yuan2020interpreting, dabkowski2017real,chen2018learning}, and studying the meaning of hidden layers~\cite{yuan2019Interpreting, olah2018the,du2018towards}, etc. In contrast, the explainability of deep graph models is still less explored. Since graph data contain limited locality information but have important structural information, it is usually not natural to directly extend image or text based methods to graphs. Recently, several techniques have been proposed to explain GNNs, such as GNNExplainer~\cite{gnnexplainer}, PGExplainer~\cite{pgexplainer}, RCExplainer~\cite{wang2022reinforced}, etc. These methods mainly focus on explaining graph nodes, edges, features, or subgraphs.

\begin{figure}
    \centering
    \resizebox{0.9\linewidth}{!}{
    \includegraphics{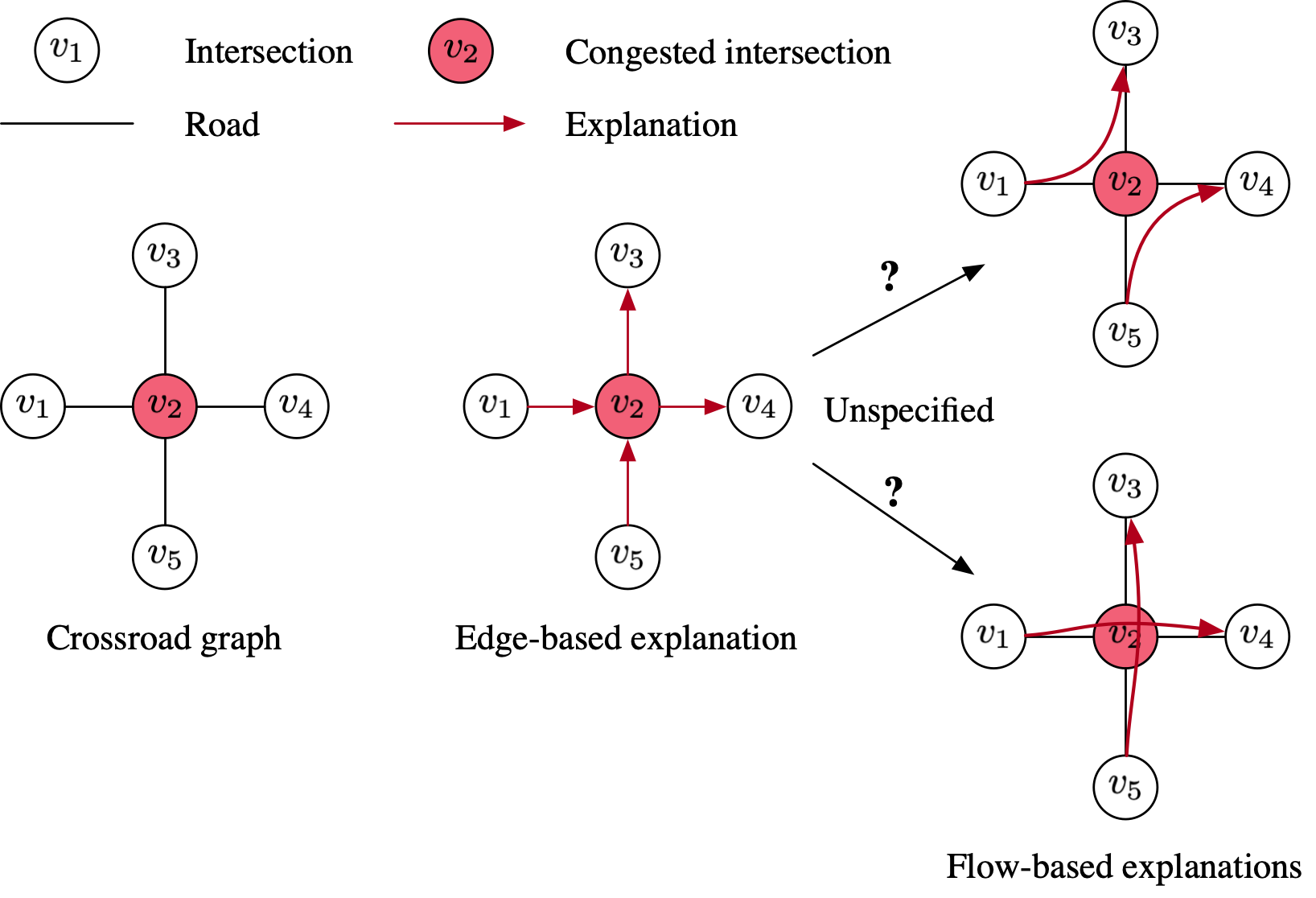}}
    \caption{\textbf{Edge-based explanation challenge.} The graph in the figure describes a traffic graph where each node indicates an intersection; each edge denotes a road. The task is to explain why there is a traffic jam at the crossroad and how to construct overpasses to address the problem, where we denote the congested intersection as a red node. The edge-based explanation indicates that the in-flow from $v_1, v_5$ to $v_2$ and the out-flow from $v_2$ to $v_3, v_4$ cause the traffic jam. However, this explanation falls short when it comes to providing the fine-grained traffic flow data necessary for overpass construction planning. In such case, the edge-based explanation is unspecified since it gives rise to two equally valid, but mutually contradictory, flow interpretations that imply conflicting overpass construction strategies, \ie, $v_1\rightarrow v_2 \rightarrow v_3$ or $v_1 \rightarrow v_2 \rightarrow v_4$.}
    \label{fig:challenge}
\end{figure}

In this study, we posit that message flows serve as the intrinsic operational mechanism of Graph Neural Networks (GNNs), offering a more intuitive and natural lens through which to investigate their explainability. Specifically, traditional edge-based explanation approaches encounter limitations when required to account for continuous messages or multi-hop correlations, which cannot be adequately represented by individual, discrete edges, as illustrated in Fig.~\ref{fig:challenge}. In this challenge, we need a fine-grained explanation to build a new overpass to solve traffic jams, but edge explanations fail to provide such information. To address this obstacle, we introduce a suite of message flow-based explanation techniques, with the primary methodology designated as FlowX. FlowX attributes GNN predictions to message flows and studies the importance of different message flows. We first develop a systematic framework that incorporates flow-based graph modeling as a foundational layer for message flows. With our framework, FlowX begins with quantifying the importance of flows by following the principle of Shapley value~\cite{shapley}, conceptualizing flows as collaborative players that collectively contribute to the model predictions. Given that message flows cannot be straightforwardly quantified for marginal contributions, we propose a fair flow sampling scheme as the initial assessment of different flows. Building upon this initial assessment, we propose a coalition-level learning-based algorithm with information sparsity controls to tailor the flow explanations serving different model explanation requirements. Our experimental evaluations encompass a range of tests: necessary and sufficient explanations~\cite{amara2022graphframex} comparisons, target edge/flow retrieval ability test, mutual information training comparisons between flow-based and edge-based methods, and visualizations. Experimental results on necessary and sufficient explanations~\cite{amara2022graphframex} show that our proposed FlowX outperforms existing methods significantly and consistently from multiple perspectives. Additional experiments validate the advantages of flow-based methods compared with other approaches. Both quantitative and qualitative studies demonstrate that the proposed FlowX and its variants lead to novel and improved explanations of GNNs.

\section{Preliminaries \& Related Work}

% \subsection{Notations}

% We provide notation definitions in Table~\ref{tab:notations} for clarity.

\subsection{Graph Neural Networks}
With the advances of deep learning, several graph neural network approaches have been proposed to solve graph tasks, including graph convolutional networks (GCNs)~\cite{gcn}, graph attention networks (GATs)~\cite{gat}, and graph isomorphism networks (GINs)~\cite{gin}, etc. They generally follow a message-passing framework to learn graph node features. Specifically, the new features of a target node are learned by aggregating message flows passed from its neighboring nodes. 
Without loss of generality, we consider the input graph as a directed graph with $n$ nodes and $m$ edges. The graph is denoted as $\mathcal{G} = (V, E)$, where $V=\{v_1, \ldots, v_{n}\}$ denotes nodes, and $E=\{e_{ij}\}$ represents edges in which $e_{ij}$ is the directed edge $v_i\rightarrow v_j$.
Then it can be represented by a feature matrix $X \in \mathbb{R}^{d \times n}$  and an adjacency matrix $A \in \mathbb{R}^{n \times n}$.
Each node $v_i$ is associated with a $d$-dimensional feature vector $\mathbf{x}_i$ corresponding to the $i$-th column of $X$.
The element $a_{ij}$ in $A$ represents the weight of $e_{ij}$, and $a_{ij}=0$ indicates $e_{ij}$ does not exist. For the $t$-th layer in GNNs, the message aggregation procedures can be mathematically written as a two-step computation as 
\begin{align}
         \mbox{Aggregate:}& \,\,S^{t} = X^{t-1}\hat{A}^t,\label{equ:MP1}\\
         \mbox{Combine:}&\,\, X^{t} = \mbox{M}^{t}(S^{t}), \label{equ:MP2}
\end{align}
where $X^{t} \in \mathbb{R}^{d_{t}\times n} $ denotes the node feature matrix computed by the $t$-th GNN layer and $X^0 = X$. Here $\mbox{M}^t(\cdot)$ denotes the node feature transformation function at layer $t$ and $\hat{A}^t$ is the connectivity matrix at layer $t$. Note that we name the elements in $\hat{A}^t$ as layer edges and $\hat{a}^t_{jk}$ indicates the layer edge connecting node $v_j$ and $v_k$ in layer $t$. 
For example, in GCNs, the transformations are defined as $\sigma(W^t S^{t})$ and $\hat{A}^t=D^{-\frac{1}{2}}(A+I)D^{-\frac{1}{2}}$ 
where $W^t\in \mathbb{R}^{d_{t} \times d_{t-1}}$ is a trainable weight matrix, $\sigma(\cdot)$ denotes the activation function, $I$ is an identity matrix to add self-loops to the adjacency matrix, and $D$ denotes the diagonal node degree matrix. We can stack $T$ GNN layers on top of each other to form a $T$-layer network, and the network function can be expressed as
\begin{equation*}\label{equ:GNN}
	f(\mathcal{G}) = g(M^T(M^{T-1}(\cdots M^{1}(X^0\hat{A}^1)\cdots )\hat{A}^{T-1})\hat{A}^T).
% 	% \vspace{10pt}
\end{equation*}
When $f(\mathcal{G})$ is a graph classification model, $g(\cdot)$ generally consists of a readout function, such as global mean pooling, and a multi-layer perceptron (MLP) graph classifier. Meanwhile, when $f(\mathcal{G})$ is a node classification model, $g(\cdot)$ represents an MLP node classifier.

\subsection{Explainability of Graph Neural Networks}
A major limitation of GNNs is their lack of explainability. Thus, different methods have been proposed to explain the predictions of GNNs, such as GraphLime~\cite{graphlime}, GNNExplainer~\cite{gnnexplainer}, PGExplainer~\cite{pgexplainer}, PGMExplainer~\cite{vu2020pgm}, SubgraphX~\cite{yuan2021explainability}, TAGE~\cite{xie2022task}, XGNN~\cite{xgnn}, GraphSVX~\cite{duval2021graphsvx}, Refine~\cite{wang2021towards}, RGExplainer~\cite{shan2021reinforcement}, and VGIB~\cite{yu2022improving}. These methods can be mainly grouped into the following categories based on the views of their explanations. First, several techniques provide explanations by identifying important nodes in the input graph.  For example, GradCAM~\cite{pope2019explainability} measures node importance by combining the hidden features and gradients; LRP~\cite{baldassarre2019explainability} and Excitation BP~\cite{pope2019explainability} decompose the predictions into several terms and assign these terms to different nodes; PGM-Explainer~\cite{vu2020pgm} builds a probabilistic graphical model by randomly perturbing the node features and employs an interpretable Bayesian network to generate explanations. GraphSVX~\cite{duval2021graphsvx} focuses on node and node feature explanations using Shapley value. GraphLime~\cite{graphlime} only targets node feature explanations using a surrogate model with a kernel-based feature selection algorithm. Second, several existing methods, such as GNNExplainer~\cite{gnnexplainer}, PGExplainer~\cite{pgexplainer}, and GraphMask~\cite{graphmask}, explain GNNs by studying the importance of different graph edges. These methods follow a similar high-level idea that learns masks to identify important edges while maximizing the mutual information. In order to achieve task-agnostic explanations, TAGE~\cite{xie2022task} targets explaining GNNs by searching an edge-based subgraph to align original graph embeddings regardless of downstream classifiers. Third, SubgraphX~\cite{yuan2021explainability} proposes to explain GNNs via subgraphs. It incorporates the Monte Carlo tree search algorithm to explore subgraphs and employs Shapley values to measure the importance. Fourth, XGNN~\cite{xgnn} focuses on model-level explanations by generating graph patterns that can maximize a certain model prediction. In addition, Refine~\cite{wang2021towards} targets explaining from both class and instance levels. RGExplainer~\cite{shan2021reinforcement} focuses on generative reinforcement learning explanations. VGIB~\cite{yu2022improving} employs the recent information bottleneck technique but it is intrinsically an interpretable method, instead of an explainer given fixed models.

Furthermore, as the emergence of causality works~\cite{peters2016causal, ahuja2021invariance, chen2022ciga, gui2023joint, gui2022good, li2023graph}, causal-based explanations including counterfactual explanations also become a valuable avenue~\cite{guo2023counterfactual}. Among these methods, \cite{bajaj2021robust} and CLEAR~\cite{ma2022clear} focus on counterfactual explanations where CLEAR can generate new counterfactual edges that do not exist in the original graph. RC-Explainer~\cite{wang2022reinforced} combines both a causality principle and reinforcement learning to search for edge explanations and serves as a new interesting and comprehensive causal baseline. While these methods explain GNNs from different views, they cannot provide explanations to solve the unspecified challenge shown in Fig.~\ref{fig:challenge}.

In prior work, GNN-LRP~\cite{gnnlrp} employs LRP with respect to graph adjacency matrices, resulting in explanations referred to as relevant walks. However, these gradient-based methods follow the $\mbox{Gradient} \times \mbox{Input}$ scheme that fails the model parameter randomization test and might not be sensitive to model parameters~\cite{adebayo2018sanity}. Since relevant walks are highly inherent to this gradient-based method, their explanation potentials are largely constrained. This limitation prompts us to introduce a higher-level abstraction, termed message flows, which broadens the explanation scope from subgraphs, nodes, and edges to flows. In Sec.~\ref{sec:causal}, we compare the explanation modelings through the lens of causality. Within our flow-based graph modeling framework, the concept of a relevant walk serves merely as a specific instance of our more expansive flow explanations. Furthermore, based on our flow framework, GNN-LRP, along with the proposed FlowX, FlowX$_{nec}$, FlowX$_{suf}$, and FlowMask offer a comprehensive suite of flow-based explanations that includes gradient-based, Shapley-based, surrogate, and hybrid methods.

% Meanwhile, . Furthermore, which will be interpreted as a special case of flows in our paper. GNN-LRP, however, follows
% A relevant walk is defined as a $T$-length ordered edge sequence that corresponds to a $T$-step directed path on the input graph.
% To study walk explainability, GNN-LRP considers the GNN prediction as a function and decomposes it using higher-order Taylor expansions to distribute prediction scores to relevant walks. Specifically, by using $T$-order Taylor expansion with a proper root, each term in the Taylor expansion corresponds to a relevant walk and is regarded as the importance score. While our proposed FlowX shares a similar explanation target, i.e., flow/walk, with GNN-LRP, our method is fundamentally different. The GNN-LRP is developed based on score decomposition while our method follows the philosophy of Shapley values from cooperative game theory as initial assessments and proposes a learning-based algorithm for the score generation. In addition, GNN-LRP has several constraints on the activation function and bias term used in GNNs while our method can be applied to general GNN models. Furthermore, as the GNN-LRP follows the $\mbox{Gradient} \times \mbox{Input}$ scheme, it may not pass the model parameter randomization test and may not be sensitive to model parameters~\cite{adebayo2018sanity}. 

\noindent{\bf Differences with Related Methods.} While Shapley value has been studied under a node-level method GraphSVX~\cite{duval2021graphsvx} and a subgraph-level method SubgraphX~\cite{yuan2021explainability}, it has never been applied under a flow level. Applying Shapley value to flows introduces different challenges and insights, \eg, flows cannot be extracted directly to serve as individual players in marginal contribution calculations (Sec.~\ref{sec:challenge1}). Moreover, there are two distinctions between the sampling processes of SubgraphX and FlowX. First, the sampling process in SubgraphX is a tree search process with a pruning strategy, while the sampling process in this work is a fair permutation-based sampling process as described in Sec.~\ref{sec:challenge2}. Second, because of the pruning strategy, the sampling process in SubgraphX is sequential. In contrast, our flow sampling process is nearly fully paralleled. Specifically, with enough GPU memory, we only require one pass model forward calculation for each graph. Furthermore, in addition to the sampling process, we propose an innovative Shapley coalition-level training tailoring the pure Shapley value to diverse explanation targets: necessary or sufficient explanations~\cite{amara2022graphframex}, which makes our method unique.

% In particular, a recent study proposes a surrogate method, known as GraphSVX~\cite{duval2021graphsvx}. However, GraphSVX is not fairly comparable with our method because it explains GNNs at levels of nodes and their features, but we explain GNNs at flow/edge level.
% Another recent study proposes SubgraphX~\cite{yuan2021explainability}, which employs a search algorithm to explore and identify subgraphs with high Shapley scores.
% While these methods use Shapley values, there are several fundamental differences. First, our proposed method focuses on explaining message flows, which are the most fine-grained units for explanations as GNNs are based on message passing schemes. Please refer to supplementary material~1 and 3.3 for more reasons of using message flow explanations. Second, we use Shapley-like values as importance score approximations, but we also provide a possible training refinement method for special importance ranking training. Due to these differences, we cannot regard these methods as similar or comparable.

\section{The Proposed FlowX} \label{proposed_method}
While existing methods mainly focus on explaining GNNs with graph nodes, edges, or subgraphs, we propose to study the explainability of GNNs from the view of message flows. We argue that message flows are the fundamental building blocks of GNNs and it is natural to study their contributions towards GNN predictions. With our message flow framework, we propose a novel method, known as FlowX, to investigate the importance of different message flows. Specifically, we follow the philosophy of Shapley values~\cite{shapley} from game theory and propose a marginal contribution approximation scheme for them. In addition, a learning-based algorithm is proposed to improve the explainability of message flows.   

\subsection{A Message Flow View of GNNs}
We consider a deep graph model with $T$ GNN layers. Each GNN layer aggregates 1-hop neighboring information to learn new node embeddings. Hence, for any node, the outgoing messages are transmitted within its $T$-hop neighbors. Then the outputs of GNNs can be regarded as a function of such transmitted $T$-step messages, which are named as message flows in this work. 
Formally, we introduce the concept of message flows and message carriers as follows:

\subsubsection{Definition 1: Message Carrier}

We use the connectivity matrix to represent the carriers for message flows. Given the connectivity matrix $\hat{A}^t$ at layer $t$, the layer edge $\hat{a}^t_{ij}$ in this matrix represents the message carrier with which the message passes from node $v_i$ to $v_j$ at layer $t$. 

Note that we use the superscript $t$ to distinguish the message carriers in different layers since their corresponding message flows are different. Then the set of all message carriers, i.e., all layer edges, is defined as $\mathbfcal{A}=\{\cdots,\hat{a}^1_{uv},\cdots, \hat{a}^t_{uv} ,\cdots,\hat{a}^T_{uv},\cdots\}$ and $|\mathbfcal{A}|=|E|\times T$.

\subsubsection{Definition 2: Message Flow}

In a $T$-layer GNN model, we use $ \mathcal{F}_{ijk\ldots \ell m}$ to denote the message flow that starts from node $v_i$ in the input layer, and sequentially passes the message to node $v_j, v_k, \dots, v_\ell$ until to node $v_m$ in the final layer $T$. The corresponding message carriers can be represented as $ \{\hat{a}^1_{ij}, \hat{a}^2_{jk}, \ldots, \hat{a}^T_{\ell m}\}$. 

\noindent{\bf Notation of flows:} In a $T$-layer GNN model, all message flows start from the input layer and end with the final layer so that their lengths are equal to $T$. For ease of notation, we introduce the wildcard "$*$" to represent any valid node sequence and $\mathbfcal{F}$ to denote the message flow set. For example, we can use $\mathbfcal{F}_{ij*}$, $\mathbfcal{F}_{*\ell m}$, and $\mathbfcal{F}_{ij * \ell m}$ to denote the message flow sets that share the same message carrier $\hat{a}^1_{ij}$, $\hat{a}^T_{\ell m}$, and both of them, respectively. In addition, we employ another wildcard "$?$" to denote any single node and $?\{t\}$ to represent any valid node sequence with $t$ nodes. For example, $\mathbfcal{F}_{?\{3\}}$ means the set of valid 2-step message flow with 3 nodes. 
Note that the following property of message flow sets also holds:
\begin{equation}\label{eq:intersection_property}
	\mathbfcal{F}_{ij * \ell m} = \mathbfcal{F}_{ij *} \cap \mathbfcal{F}_{ * \ell m}.
\end{equation}

Intuitively, Property~\ref{eq:intersection_property} indicates that given two sets of flows where the first set includes all flows \textit{starting} from $v_i\rightarrow v_j$, and the second set contains all flows \textit{ending} with $v_\ell \rightarrow v_m$, the intersection flow set of these two flow sets consists of all flows \textit{starting} from $v_i\rightarrow v_j$ \textit{and ending} with $v_\ell \rightarrow v_m$.

The final embedding of node $v_m$ is determined by all incoming message flows to node $v_m$, which can be denoted as $\mathbfcal{F}_{*m}$. Since the output of the GNN model is obtained based on the final node embeddings, it is reasonable to treat the GNN output as the combination of different message flows. Hence, it is natural to explain GNN models by studying the importance of different message flows towards GNN predictions. 

\subsection{Flow-based Graph Modeling}\label{sec:causal}

\begin{figure}
    \centering
    \resizebox{1\linewidth}{!}{
    \includegraphics{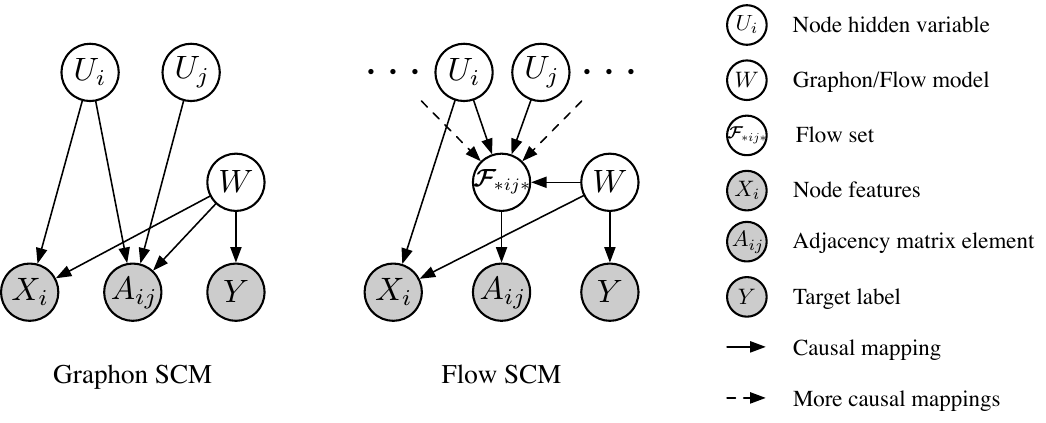}}
    \caption{\textbf{Graph modeling using SCMs.} The figure illustrates the assumptions of data generation processes described by Structural Causal Models (SCMs). The grey nodes are observed and can be intervened by us. The white nodes are unobserved or cannot be intervened/operated directly.}
    \label{fig:SCMs}
\end{figure}

To intuitively demonstrate the correlations among node-based, edge-based, subgraph-based, and flow-based explanations, we propose to model graph generation processes through the lens of causality~\cite{peters2016causal, pearl2009causality}. Generally, the commonly used graphon model~\cite{lovasz2006limits, bevilacqua2021size} $W$ is a measurable function modeling the relations/edges between nodes, \ie, it maps two node hidden variables $U_i$ and $U_j$ to one adjacency matrix element $A_{ij}$, as shown in Fig.~\ref{fig:SCMs}. When dealing with identical nodes, the graphon model $W$ produces the observable node signals/features $X_i$. Acting as a confounder, $W$ creates explainable correlations between $X_i$/$A_{ij}$ and $Y$, enabling us to use observed variables $X_i$ and $A_{ij}$ to explain the model prediction $Y$. Depending on the variables employed, explanations can fall into categories such as node-based/node-feature-based ($X_i$), edge-based ($A_{ij}$), or subgraph-based (hybrid). 

However, the inherent limitation of the graphon function $W$ lies in its capacity to model only pairwise relations between nodes, \ie, edges. To explain messages that transmit across multiple nodes, we extend the $W$ to model the relations of multiple nodes in sequences. For a $T$-hop message passing network, $W$ maps $T+1$ input node hidden variables to one message flow. Specifically, in Fig.~\ref{fig:SCMs}, the flow model $W$ associates node sequences, including $U_i$ and $U_j$, to the set of flows $\mathbfcal{F}_{*ij*}$. Then, the relations between two nodes $A_{ij}$ can be modeled by the combinations of these message flows $\mathbfcal{F}_{*ij*}$. Note that, we abuse the notation $W$ only in this subsection to maintain consistency with the previous work~\cite{bevilacqua2021size}.

\subsection{Marginal Contribution Calculation by Flow Samplings}
While explaining GNNs with message flows seems to be promising, it is still crucial to properly measure the importance of those message flows. Our FlowX proposes to follow the philosophy of Shapley value~\cite{shapley} to use the marginal contributions in different flow sets as the initial assessments of flow importance. Shapley value is a solution concept in cooperative game theory and is used to fairly assign the game gain to different players. When considering marginal contributions in GNN explanation tasks, we treat the message flows as different players and the GNN prediction score as the total game gain. Formally, given the trained GNN model $f(\cdot)$ and the input graph $\mathcal{G}$, we use $\mathbfcal{F}_{*}$ to denote the set of all valid flows, \textit{i.e.}, all players in the game. Then given any flow $\mathcal{F}^k$, we mathematically define the contribution as

\begin{equation}
\phi(\mathcal{F}^k)=\sum_{P\subseteq \mathbfcal{F}_{*} \setminus\{\mathcal{F}^k\}}\mathcal{W}(|P|)( f\left(P\cup\{\mathcal{F}^k\}\right)-f(P) ),\label{eq:shap} 
\end{equation}
where $\phi(\cdot)$ denotes the flow score of $\mathcal{F}^k$; $P$ denotes the possible coalition group/set of players; $|\cdot|$ is the set size; $\mathcal{W}(\cdot)$ is a weight function assigned according to the size of the group $P$.
Here $f\left(P\cup\{\mathcal{F}^k\}\right)-f(P) $ is the marginal contribution of flow $\mathcal{F}^k$ for a particular coalition group, which can be computed by the prediction difference between combining $\mathcal{F}^k$ with the coalition group $P$ and only using $P$. Note that Eq.~\ref{eq:shap} is equivalent to the classic Shapley value when $\mathcal{W}(|P|)=\frac{|P|!\left(|\mathbfcal{F}_{*}|-|P|-1\right)!}{|\mathbfcal{F}_{*}|!}$ where $!$ is the factorial operator.
To compute the flow score  $\phi(\mathcal{F}^k)$, we need to enumerate all possible coalition groups and considers different interactions among players. 
% However, it is time-consuming to consider all possible coalition; thus, we sample several marginal contributions to approximate the final flow score.
%\textcolor{blue}{Note that Eq.~\ref{eq:shap} is a fair solution since it is the only solution satisfying four desirable properties, including efficiency, symmetry, linearity, and dummy~\cite{shap}.(this claim is no longer valid.)} 
However, calculating flow Shapley values introduces two challenges.

\subsubsection{Challenge 1: Indirect Intervention Targets}\label{sec:challenge1}

\begin{figure*}[t!]
    \centering
    % \vspace{-0.8cm}
    \includegraphics[width=0.95\linewidth]{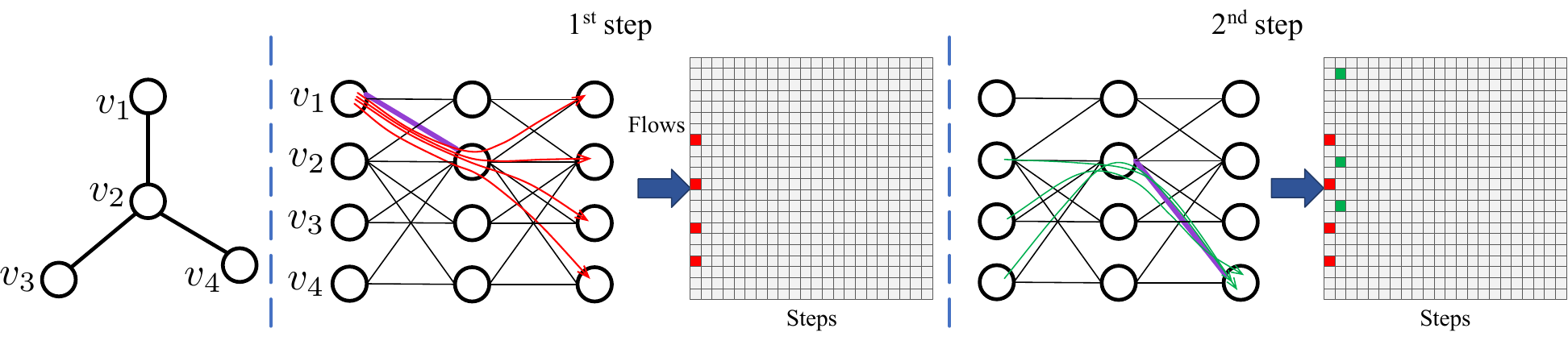}
    \vspace{-0.6cm}
    \caption{An illustration of our initial assessments via sampling marginal contributions. For each sampling iteration, we iteratively remove one layer edge until all layer edges are removed. In this example, the removed layer edges are shown in bold and purple lines while the corresponding message flows are shown in arrow lines. In the first step, we remove the layer edge between $v_1$ and $v_2$ from the first GNN layer and compute the marginal contribution. Then four message flows (red) are removed and the contribution scores are averaged and assigned to these four message flows. In the second step, we remove the layer edge $\hat{a}^2_{24}$ and distribute the marginal contribution to the corresponding three flows (green).}
    % \vspace{-0.2cm}
    \label{fig:flow_sampling}
\end{figure*}

The first challenge is message flows cannot be removed/intervened independently from both the input level and the model level, which is also implied by Fig.~\ref{fig:SCMs}. Therefore, it is impossible to directly compute their importance scores using Eq. (\ref{eq:shap}). To overcome this challenge, we consider the finest component we can intervene (remove) to calculate marginal contributions, \ie, layer edges.
For example, since the flow set $\mathbfcal{F}_{ij*}$ depends on layer edge $\hat{a}^1_{ij}$, when we remove $\hat{a}^1_{ij}$, all flows in flow set $\mathbfcal{F}_{ij*}$ will be simultaneously removed from the model. Fig.~\ref{fig:flow_sampling} illustrates another example, where removing layer edge $\hat{a}^1_{12}$ is equivalent to removing flows $\mathbfcal{F}_{12?\{1\}}=\{\mathcal{F}_{121}, \mathcal{F}_{122}, \mathcal{F}_{123}, \mathcal{F}_{124}\}$. In this scenario, since multiple flows are removed at the same time, the obtained marginal contributions should be distributed to these flows, \eg, even distributions.

\subsubsection{Challenge 2: Fair Flow Sampling}\label{sec:challenge2}

Since enumerating all possible coalition groups is time-consuming especially when the input graph is large-scale and the GNN model is deep, a sampling process is required. However, applying simple Monte Carlo (MC) sampling~\cite{vstrumbelj2014explaining} cannot guarantee a fair sampling for each flow. Therefore, as shown in Algorithm~\ref{al:1}, we propose an innovative flow sampling scheme that divides the sampling process into two levels and ensures each flow can be and only be sampled one time per iteration, leading to a fair flow sampling process.

\begin{algorithm*}[ht!]
    \caption{ \textsc{Initial approximations of flow importance scores.}}
    \label{al:1}
    \begin{algorithmic}[1]
        \STATE  Given a trained GNN model $f(\cdot)$ and an input graph, the set of all layer edges is represented by $\mathbfcal{A}$. For each message flow $\mathcal{F}^k$, two $|\mathbfcal{A}|$-dimensional vectors $\mathbf{s}^{\mathcal{F}^k}$ and $\mathbf{c}^{\mathcal{F}^k}$ denote its importance scores and removing index counts respectively. In addition, $\widehat{\mathbfcal{A}}$ denotes the set of removed layer edges and $\widehat{\mathbfcal{F}}$ is the set of removed message flows.
        \STATE  For each flow $\mathcal{F}^k$, initialize  $\mathbf{s}^{\mathcal{F}^k}$ and $\mathbf{c}^{\mathcal{F}^k}$ as zero vectors. 
        
        \FOR{iteration $i$ from 1 to flow sampling iteration $M$}
        
        \STATE Initialize the removed sets as empty that $\widehat{\mathbfcal{A}} = \emptyset$ and $\widehat{\mathbfcal{F}} = \emptyset$.
        
        \STATE Randomly shuffle and permute the layer edge set $\mathbfcal{A}$, denoted as $\mathbfcal{A}^{\bm{\pi}^i}$.
        
        \FOR{$j$ from 1 to $|\mathbfcal{A}|$}
        
        \STATE Select the $j$-th layer edge in $\mathbfcal{A}^{\bm{\pi}^i}$, denoted as $\hat{a}^t_{\ell m}$.
        
        \STATE Block the layer edge  $\hat{a}^t_{\ell m}$ in GNN model $f(\cdot)$, then the removed flows are $\widehat{\mathbfcal{F}}^{j} = \mathbfcal{F}_{?\{t-1\}\ell m?\{T-t\}} \setminus(\widehat{\mathbfcal{F}}\cap \mathbfcal{F}_{?\{t-1\}\ell m?\{T-t\}})$.
        
        \STATE Compute the prediction difference that  $s_j = f(\mathbfcal{A}\setminus\widehat{\mathbfcal{A}}) - f(\mathbfcal{A}\setminus(\widehat{\mathbfcal{A}}\cup\hat{a}^t_{\ell m}))$.
        
        \STATE Update $\widehat{\mathbfcal{A}} = \widehat{\mathbfcal{A}}\cup\hat{a}^t_{\ell m}$ and $\widehat{\mathbfcal{F}}= \widehat{\mathbfcal{F}}\cup \mathbfcal{F}_{?\{t-1\}\ell m?\{T-t\}}$.
        
        \STATE Compute averaged score that $\bar{s_j}= s_j/|\widehat{\mathbfcal{F}}^{j}|$. 
        
        \FOR{each flow $\mathcal{F}^k$ in $\widehat{\mathbfcal{F}}^{j}$}
        \STATE Update $s^{\mathcal{F}^k}_j = s^{\mathcal{F}^k}_j+\bar{s_j} $.
        
        \STATE Update $c^{\mathcal{F}^k}_j = c^{\mathcal{F}^k}_j + 1 $
        \ENDFOR
        \ENDFOR
        \ENDFOR
        \STATE For each message flow $\mathcal{F}^k$, compute the marginal contribution vector  $\mathbf{s}^{\mathcal{F}^k} =\mathbf{s}^{\mathcal{F}^k}/\mathbf{c}^{\mathcal{F}^k}  $
    \end{algorithmic}
\end{algorithm*}

\subsubsection{Flow Sampling}

Formally, let $M$ denote the total number of sampling iterations. In the $i$-th sampling iteration, we generate a random permutation vector $\bm{\pi}^i$ of $|\mathbfcal{A}| = |E|\times T$ elements. Then all layer edges in $\mathbfcal{A}$ will be permuted based on $\bm{\pi}^i$, denoted as $\mathbfcal{A}^{\bm{\pi}^i}$ that $\mathbfcal{A}^{\bm{\pi}^i}_j = \mathbfcal{A}_{\bm{\pi}^i_j}$. Subsequently, in each iteration, there are $|\mathbfcal{A}^{\bm{\pi}^i}|$ steps. In each step, we remove one layer edge from $\mathbfcal{A}^{\bm{\pi}^i}$ in order, and compute the marginal contributions of corresponding removed flows as shown in Fig.~\ref{fig:flow_sampling}. We use $\widehat{\mathbfcal{A}}$ to denote the set of layer edges that have been removed in the current iteration and $\widehat{\mathbfcal{F}}$ to denote the set of corresponding removed message flows. These two sets are initialized as $\widehat{\mathbfcal{A}} = \emptyset$ and $\widehat{\mathbfcal{F}} = \emptyset$ in the beginning of each sampling iteration. 
Specifically, at step $j$ in the current iteration, the $j$-th element of $\mathbfcal{A}^{\bm{\pi}^i}$ is removed. Assuming the removed layer edge is $\hat{a}^t_{\ell m}$, the computation operations can be mathematically written as
\begin{align}
s_j &= f(\mathbfcal{A}\setminus\widehat{\mathbfcal{A}}) - f(\mathbfcal{A}\setminus(\widehat{\mathbfcal{A}}\cup\hat{a}^t_{\ell m})), \\
\widehat{\mathbfcal{A}} &= \widehat{\mathbfcal{A}}\cup \{\hat{a}^t_{\ell m}\},\\
\widehat{\mathbfcal{F}}^{j} &= \mathbfcal{F}_{?\{t-1\}\ell m?\{T-t\}} \setminus(\widehat{\mathbfcal{F}}\cap \mathbfcal{F}_{?\{t-1\}\ell m?\{T-t\}}),\\
\widehat{\mathbfcal{F}} &= \widehat{\mathbfcal{F}}\cup \mathbfcal{F}_{?\{t-1\}\ell m?\{T-t\}},
\end{align}
where $\widehat{\mathbfcal{F}}^{j}$ denotes the removed message flows at step $j$ by removing $\hat{a}^t_{\ell m}$; similar to the set difference operation, "$\setminus$" denotes the removal of layer edges or flows. Note that $\widehat{\mathbfcal{F}}^{j}$ is not equivalent to $\mathbfcal{F}_{?\{t-1\}\ell m?\{T-t\}}$ since parts of the flows in $\mathbfcal{F}_{?\{t-1\}\ell m?\{T-t\}}$ may have been already removed in the previous steps. Given these removed flows $\widehat{\mathbfcal{F}}^{j}$, the score is averaged by the number of flows, \ie, $\bar{s_j}=s_j/|\widehat{\mathbfcal{F}}^{j}|$, and we distribute $\bar{s_j}$ to each flow in $\widehat{\mathbfcal{F}}^{j}$. By repeating such operations until all layer edges are removed/calculated, each flow ends up obtaining one distributed score. 
% we obtain $|\mathbfcal{A}|$ marginal scores and assign them to the corresponding flows. 
Note that the order information in $\mathbfcal{A}^{\bm{\pi}^i}$ is important since in the earlier 
steps, the removed flows $\widehat{\mathbfcal{F}}^{j}$ are interacting with a larger coalition group $\mathbfcal{A}\setminus(\widehat{\mathbfcal{A}}\cup\hat{a}^t_{\ell m})$, while in the later steps the coalition groups are generally smaller. Altogether, the steps of our marginal contribution sampling are shown in Algorithm~\ref{al:1}.

After $M$ flow sampling iterations, our method explores $M$ permutations of layer edge set and each flow is sampled to obtain $M$ marginal contributions. 
For each flow $\mathcal{F}^k$, we use a $|\mathbfcal{A}|$-dimensional vector to store its marginal contributions, denoted as $\mathbf{s}^{\mathcal{F}^k}$, where $\mathbf{s}^{\mathcal{F}^k}_j$ is the importance score obtained when $\mathcal{F}^k$ is removed at step $j$. To obtain $\mathbf{s}^{\mathcal{F}^k}$, valid scores at the same step and different sampling iterations are averaged, where a score of a flow is valid at step $j$ if the flow is removed at this step. The concrete operation is shown in line 18 of the algorithm~\ref{al:1}, where "/" denotes an element-wise division. To avoid zero-divisions, we add a small value $10^{-10}$ to every element in $\mathbf{c}^{\mathcal{F}^k}$. The importance score will be a flow Shapley value approximation if we compute the summation of the elements in $\mathbf{s}^{\mathcal{F}^k}$.
% For example, if a flow is removed multiple times at step $j$ in different sampling iterations, the scores are averaged, which indicates scores obtained from coalitions at the same step and different iterations are weighed equally.

% Then it raises the question that how to convert $\mathbf{s}^{\mathcal{F}^k}$ to the final importance score of $\mathcal{F}^k$. The final importance score will be a flow Shapley value approximation if we simply compute the summation of the elements in $\mathbf{s}^{\mathcal{F}^k}$. While Shapley values are promising, there are different types of explanations we can achieve, \eg, counterfactual explanations~\cite{guo2023counterfactual}, or sufficient explanations that maintain model performances.

% the approximations might not be accurate since $M$ is always set as $M<<|\mathbfcal{A}|!$ for the sake of computation efficiency. Moreover, the importance of the early steps when more layer edges exist in the model should be different to the importance of later steps when only few layer edges exist in diverse situations. Hence, directly computing the summation of the scores in $\mathbf{s}^{\mathcal{F}^k}$ is not appropriate. To overcome this issue, we propose the following training refinement method.

\subsection{Trainable FlowX for Diverse Explanations}\label{learning}
While Shapley values are promising, there are different types of explanations we can achieve, \eg, counterfactual explanations~\cite{guo2023counterfactual}, or sufficient explanations that maintain model performances. Making an explainer trainable can not only refine importance scores but also enable the explainer to serve different purposes, \eg, searching for necessary components or sufficient components. These two aspects are firstly proposed by~\cite{survey} and further characterized by~\cite{amara2022graphframex}. Specifically, we relate selecting necessary components as a process of finding counterfactual subgraphs that can produce the most significant different performance. In this process, the complementary subgraphs of the counterfactual subgraphs are necessary components of the predictions. On the other hand, selecting sufficient components is the process of finding subgraphs that can maintain similar or even better model performances.

Different from optimizing classification tasks, training such an explainer is intrinsically a harder task as described in Sec.~\ref{sec:information}. Specifically, with our obtained score vectors $\mathbf{s}^{\mathcal{F}^k}$, we propose to consider these values as initial assessments of flow importance and learn associated importance scores by injecting randomness. In this training process, instead of retraining individual flow scores, we only learn weights of coalitions $\mathbf{w}$ that are used to sum the elements in $\mathbf{s}^{\mathcal{F}^k}$ to obtain the final importance score for each flow. Note that learning $\mathbf{w}$ can be considered as learning $W(P)$ in Eq.~\ref{eq:shap}.
% Note that since message flows cannot be masked out individually, we can only directly focus on the masks of layer edges and indirectly study the message flows. 

\subsubsection{From Flow Scores to Layer Edge Scores}

Formally, given the input graph $\mathcal{G}$ and GNN model $f(\cdot)$, we first obtain marginal contribution vector $\mathbf{s}^{\mathcal{F}^k}$ for each message flow. Then we apply a dot product between the only trainable weight vector $\mathbf{w}$ and the flow score vector $\mathbf{s}^{\mathcal{F}^k}$ to attain the final flow score $s^{\mathcal{F}^k}$. Since layer edges are the finest components we can operate in an MPNN, we convert these flow scores to layer edge scores, where summations are applied to the flows that share the same layer edges as follows. Note that the weight vector $\mathbf{w}$ is initialized by value 0.5 and is shared across all flows in the same graph, which explains our training intuition, \ie, coalitions in different steps are not equally important. These processes can be written as

% Specifically, with random noise that shares the same dimension of $\mathbf{s}^{\mathcal{F}^k}$

% Next, the flow importance is converted to layer edge importance by simply summing up the scores of all flows sharing a particular layer edge as the message carrier. Mathematically, it can be written as 
\begin{align}
s^{\mathcal{F}^k} &= \mathbf{s}^{\mathcal{F}^k}\cdot \mathbf{w} \quad \forall  \mathcal{F}^k \in \mathbfcal{F}_*, \label{eq:flow_wei_sum}\\ 
s^{\hat{a}^t_{uv}} & = \sum_{\mathcal{F}^j \in \tilde{\mathbfcal{F}}} s^{\mathcal{F}^j} \label{eq:flow_to_layer_edge}%\quad  \forall \mathcal{F}^j \quad  \textbf{with carrier} \quad  \hat{a}^t_{uv},
\end{align}
where $s^{\cdot}$ denotes the score of layer edges or flows; $\tilde{\mathbfcal{F}} = \mathbfcal{F}_{?\{t-1\}uv?\{T-t\}}$ denotes the flows that share layer edge $\hat{a}^t_{uv}$. Then based on  $s^{\hat{a}^t_{uv}}$,  we can obtain a mask vector indicating the importance of different layer edges, denoted as $\mathbf{m}$:

\begin{equation}\label{eq:sampling}
    \mathbf{m} = g(\mathbf{s}^{\mathbfcal{A}}),
\end{equation}

where $\mathbf{s}^{\mathbfcal{A}}$ denotes all layer edges' importance scores. $g(\cdot)$ is defined as an operation that sequentially includes an InstanceNorm~\cite{ulyanov2016instance} and a Gumbel-Sigmoid~\cite{jang2016categorical} to inject randomness and normalize the mask. We provide more analyses in Sec.~\ref{sec:information}.

% element-wise exponential scaling $g^e(x) = x^r$ (r is a hyper-parameter), and an output normalization. We provide more intuitions in next subsection~\ref{sec:information}.

\subsubsection{Necessary Explanations}

By applying the mask to layer edges, important layer edges are restricted and the model prediction becomes 
\begin{equation}\label{eq:nec_y}
	\hat{\mathbf{y}} = f(\mbox{Combine}(\mathcal{G}, 1-\mathbf{m})),
\end{equation}
where $\hat{\mathbf{y}}$ is the prediction vector and $\mbox{Combine}(\mathcal{G}, 1-\mathbf{m})$ denotes a modified graph where layer edges are masked out from $\mathcal{G}$ based on the values of $\mathbf{m}$. Intuitively, if important layer edges are restricted, then the prediction should change significantly. Hence, $\hat{\mathbf{y}}$ is encouraged to be different from the original prediction by learning proper weights $\mathbf{w}$. This encouragement may be also considered as searching a counterfactual graph~\cite{guo2023counterfactual}, while the distinction resides at that we focus on explanations in original graphs, but counterfactual explanations generate new graphs/edges.
Specifically, we employ the log-likelihood loss as 
\begin{equation}\label{eq:nec_loss}
\mathcal{L}(\hat{\mathbf{y}},y) = \sum_{c=1}^{|\mathcal{Y}|} \textbf{1}\{y=c\}\log \hat{y}_c,
\end{equation}
where $y\in \mathcal{Y}$ is the label representing the predicted class of the original graph $\mathcal{G}$;  $|\mathcal{Y}|$ is the number of classes;  $\textbf{1}\{\cdot\}$ denotes the indicator function, and $\hat{y}_c$ is the predicted probability for class $c$. % Furthermore, we apply exponential scaling on the layer edge mask $M$ to facilitate training as we observe that the training is not stable without such a scaling term.
% Similarly, we can optionally choose mutual information (MI)~\cite{gnnexplainer} as the loss function.

\subsubsection{Sufficient Explanations}

Considering sufficient explanations, we retain the layer edges denoted by $\mathbf{m}$ and train to maintain model performances. Therefore, the model prediction and the loss in Eq.~\ref{eq:nec_y} and \ref{eq:nec_loss} are replaced by

\begin{equation}\label{eq:suf_y}
	\hat{\mathbf{y}} = f(\mbox{Combine}(\mathcal{G}, \mathbf{m})),
\end{equation}
and
\begin{equation}\label{eq:suf_loss}
\mathcal{L}(\hat{\mathbf{y}},y) = - \sum_{c=1}^{|\mathcal{Y}|} \textbf{1}\{y=c\}\log \hat{y}_c.
\end{equation}

Note that Eq.~\ref{eq:nec_loss} can be interpreted as a counterfactual training while training negative log-likelihood (Eq.~\ref{eq:suf_loss}) can be interpreted as maximizing a lower bound of mutual information $I(Y;\mbox{Combine}(\mathcal{G}, \mathbf{m}))$ according to Eq.(6) of \cite{miao2022interpretable}. After training, the importance scores of necessary/sufficient explanations for different flows can be obtained via Eq.~\ref{eq:flow_wei_sum}. 

To distinguish the notations of different training strategies, we denote the necessary-oriented FlowX as FlowX$_{nec}$, the sufficient-oriented FlowX as FlowX$_{suf}$, and the pure flow sampling as FlowX. In addition, we introduce a training-based flow explainer variant in Sec.~\ref{sec:flowmask}, denoted as FlowMask.

\subsection{Training Dilemma \& Marginal Distribution Control}\label{sec:information}

Training importance scores is essential for a ranking/sorting process that conceptually requires pair-wise importance comparisons in simple $O(n^2)$ sorting algorithms. This problem can be considered as a regression task without direct ground-truth values, making this task more challenging than both the classification and regression tasks with explicit ground-truth. Empirically, we observe that a naive approach to this problem often results in deteriorated performance compared to the initial results obtained through Algorithm~\ref{al:1}. Therefore, we strive to reformat this problem into a binary classification task with the help of the Sparsity requirement~\ref{sec:sparsity}. This Sparsity requirement introduces a percentage threshold, $\gamma$, which effectively designates the top $1-\gamma$ layer edges as important (label 1) while relegating the remainder to the category of unimportant (label 0).
% that describes the percentage of the most important parts of graphs we are interested in. The Sparsity value $\gamma$ plays as a percentage threshold of the newly formatted binary classification problem. Specifically, the $1-\gamma$ layer edges will be selected as important (label 1), while the other $\gamma$ part will be assigned with label 0.

Given this reformatted binary classification, there are two challenges remaining, \ie, discretizing $\mathbf{m}$ and the enforcement of the selection ratio $1-\gamma$. To be specific, we tackle the first challenge by adopting a Bernoulli sampling process that incorporates stochasticity, thereby facilitating escape from local minima. The sampling process $g(\cdot)$, as outlined in Eq.~\ref{eq:sampling}, employs an InstanceNorm and a reparameterization trick Gumbel-Sigmoid~\cite{jang2016categorical} to make the process differentiable, where each value after the InstanceNorm layer are considered as the log-probability in a Bernoulli distribution. For the second challenge, we propose to combine the Sparsity and the marginal distribution control~\cite{miao2022interpretable} to set soft thresholds. Essentially, considering every element in $\mathbf{m}$ as a random variable, the soft $1-\gamma$ threshold is interpreted as a prior that every element has a $1-\gamma$ probability of being labeled as important (1), otherwise, unimportant (0). Statistically, when the length of $\mathbf{m}$ reaches infinity, the ratio of selected elements asymptotically approaches $1-\gamma$. Therefore, this marginal distribution control is a regularization that applies a penalty on the difference (KL divergence) between $\mathbf{m}$ and a vector of random variables with probability $1-\gamma$ to be 1. Formally, the regularizer $\mathcal{L}_{\gamma}$ can be written as

\begin{equation}
    \mathcal{L}_{\gamma} = \sum_{m\in \mathbf{m}} m\log{\frac{m}{1-\gamma}} + (1-m)\log{\frac{1-m}{\gamma}}.
\end{equation}
Therefore, our final optimization objective is

\begin{equation}
    \mathbf{w} = \text{arg\,min}_{\mathbf{w}} \mathcal{L}(\mathbf{y}, y) + \mathcal{L}_{\gamma}.
\end{equation}
After training, the flow scores are obtained by applying Eq.~\ref{eq:flow_wei_sum}.

\subsection{Complexity Analysis}

\subsubsection{Message Flows}

Considering the total number of message flows $|\mathbfcal{F}|$, its loose upper bound is $|\mathbfcal{F}|=O(|E|^T)$. With the consideration of connectivity, given the largest outgoing degree of nodes in the graph $d_{+}$, the tighter upper bound of the number of message flows is $|\mathbfcal{F}|=O(|E|(d_{+})^{T - 1})$.

\subsubsection{Marginal Contribution Sampling}

It is noticeable that the deep model's forward operations are the most time-consuming operations, denoted as $O(\mathcal{T}_f)$. Without any parallel consideration, the time complexity of marginal contribution sampling is $\mathcal{T}_{mcs}=O(M|\mathbfcal{A}|\mathcal{T}_f)$. When we consider parallel implementations, the most simple improvement is to use the GPU acceleration to move lines 9-15 out of the two loops and to make line 8 executed in $O(1)$, which leads to the time complexity $\mathcal{T}_{mcs} = O(M |\mathbfcal{A}| + \mathcal{T}_f)$. The space complexity should be considered because there is a big flow score matrix $\mathcal{S}_{cpu} = O(M|\mathbfcal{A}||\mathbfcal{F}|)$. As for the graphic memory requirement, this parallel implementation consumes $\mathcal{S}_{gpu} = O(M|\mathbfcal{A}|\mathcal{S}_f)$, where $O(\mathcal{S}_f) = O((|V| + |E|)dT + \mathcal{S}_{p})$ is the deep model memory complexity; here, $|V|=n$ denotes the number of nodes; $\mathcal{S}_{p}$ is the model's parameter memory complexity. In our implementation, there is a trade-off that we only parallelly calculate the model outside the innermost loop, so that our time complexity is $\mathcal{T}_{mcs}=O(M(|\mathbfcal{A}| + \mathcal{T}_f))$ and the space complexity of GPUs is $\mathcal{S}_{gpu} = O(|\mathbfcal{A}|\mathcal{S}_f)$. Generally, we set $M=50$ which is enough for stable explanations.

\section{Experimental Studies}\label{experimental_studies}

We conduct experiments to validate the effectiveness of our explainer FlowX. Specifically, we target answering the following research questions. \textbf{RQ1:} How do FlowX and FlowX$_{nec}$ perform in terms of explanations of necessity? \textbf{RQ2:} How do FlowX and FlowX$_{suf}$ perform in terms of explanations of sufficiency? \textbf{RQ3:} What are the benefits brought by flow-based methods?

\subsection{Datasets and Baselines}\label{data}

\subsubsection{Datasets}

\begin{table*}[t]
	\caption{Statistics and properties of seven datasets. Note that ``NC'' denotes node classification, and ``GC'' denotes graph classification. \# nodes denotes the number of nodes of the largest graph in the datasets for the split of explanations. Acc. represents test accuracy. Syn. denotes Synthesis datasets; while Real denotes real-world datasets;}\label{tab:datasets}
	% \small
	\centering
	\begin{tabular}{lccccc}
		\toprule
		Datasets & Task & \# graph & \# nodes & GCN Acc.  &  GIN Acc. \\
		\midrule
            BA-Infection &       Syn./GC  &         2000 & 39  &      99.00\%  &   99.50\%  \\
            BA-Traffic &    Syn./GC  &       2000 & 20 & 100.00\% & 100.00\% \\
		BA-Shapes &       Syn./NC &               1 &  700 &    90.29\% &   89.57\%  \\
		BA-LRP &        Syn./GC &             20000 & 20 &   97.95\% &   100\% \\
		ClinTox &        Real/GC &             1478 & 136 &     93.96\% &       93.96\% \\
		Tox21 &         Real/GC &            7831 & 58 &   88.66\% &    91.02\% \\
		BBBP &       Real/GC &               2039 & 100 &      87.80\% &   86.34\% \\
		BACE &       Real/GC &               1513 & 73 &      78.29\% &   80.26\% \\
		Graph-SST2 &       Real/GC &               70042 & 36 &      90.84\% &   90.91\% \\
		\bottomrule
	\end{tabular}
\end{table*}

We employ nine different datasets to demonstrate the effectiveness of our proposed FlowX with both quantitative studies and qualitative visualization results. These datasets are BA-Infection, BA-Traffic, BA-Shapes~\cite{gnnexplainer}, BA-LRP~\cite{gnnlrp}, ClinTox~\cite{moleculenet}, Tox21~\cite{moleculenet}, BBBP~\cite{moleculenet}, BACE~\cite{moleculenet}, and Graph-SST2~\cite{survey}, which include both synthetic and real-world data. Specifically, BA-Infection and BA-Traffic are datasets proposed by this work, where BA-Traffic (Sec.~\ref{sec:target_retrieval_ability}) simulates the traffic jam challenge, as shown in Fig.~\ref{fig:challenge}, to emphasize the benefits of flow-based explanations. Detailed descriptions of BA-Infection can be found in the supplementary material. BA-Shapes is a node-classification synthetic dataset that is built by attaching house-like motifs to the base Barab$\acute{\textrm{a}}$si-Albert graph where the node labels are determined by their own identifies and localizations in motifs.  Then, BA-LRP is a graph-classification synthetic dataset that includes Barab$\acute{\textrm{a}}$si-Albert graphs and the two classes are node-degree concentrated graph and evenly graph. 
Next, ClinTox, Tox21, BBBP, and BACE are real-world molecule datasets for graph classification. The chemical molecule graphs in these datasets are labeled according to their chemical properties, such as whether the molecule can penetrate a blood-brain barrier. Finally, Graph-SST2 is a natural language sentimental analysis dataset that converts text data to graphs. These graphs are labeled by their sentiment meanings. 
The properties and statistics of these datasets are included in Table~\ref{tab:datasets}.

\subsubsection{GNN Models}

In our experiments, we consider GCNs~\cite{gcn} and GINs~\cite{gin} as our graph models for all datasets. We adopt 2-layer GNNs for node classification and 3-layer GNNs for graph classification. The graph models are trained to achieve competitive performance and the details are reported in Table~\ref{tab:datasets}.

\subsubsection{Baselines}

With the trained graph models, we quantitatively and qualitatively compare our methods with ten baselines. Specifically, we provide a justification for the choice of baselines as follows. We adopt GradCAM~\cite{pope2019explainability} and DeepLIFT~\cite{deeplift} as traditional explanation baselines; GNNExplainer~\cite{gnnexplainer}, PGExplainer~\cite{pgexplainer}, PGMExplainer~\cite{vu2020pgm}, and SubgraphX~\cite{yuan2021explainability} as pioneer GNN explanation baselines. Since GNN-GI and GNN-LRP~\cite{gnnlrp} can be considered as special cases of our flow modeling~\ref{sec:causal}, they serve as flow-based explanation baselines. Considering recent new techniques, we employ RC-Explainer~\cite{wang2022reinforced} as our causal baseline and VGIB~\cite{yu2022improving} as an information bottleneck baseline. Since these baselines have different explanation targets, we set the explanation target to graph edges for fair comparisons, \ie, the explanations of different methods are converted to edge importance scores if needed. It is critical to note that GraphSVX~\cite{duval2021graphsvx} focuses on explanations of node and node features, where node feature explanations are out of the scope of our work. In addition to the loss of node feature explanations, converting its node explanations to edge explanations introduces another level of information loss, rendering the comparisons hard to be fair. Furthermore, if we consider an edge-based Shapley value method, because of the additive property of the Shapley method, edge-based Shapley values can be equivalent to the summations of finer-grained Shapley values (naive flow sampling scores), leading to an uninterest comparison when comparing under edge-based settings. More implementation details about explanation methods setting and GNN models can be found in the supplementary material~2.2. We used the datasets and implementations of the comparing algorithms in the DIG library~\cite{dig}.

\subsubsection{Flow-based Explainer Variant: FlowMask}\label{sec:flowmask}

%two studides. 1. FlowX vs no learning. 2. FlowX vs no initial assessment.

In order to make this work comprehensive, we incorporate a mutual information training based flow-based method FlowMask, which sets $s^{\mathcal{F}^k}$ in Eq.~\ref{eq:flow_wei_sum}, as a training parameter for each flow $\mathcal{F}^k$. These trainable parameters for flows form a flow mask that indicates flow importance scores. Following Eqs.~\ref{eq:flow_to_layer_edge} and \ref{eq:sampling}, we then obtain a layer edge mask. Given the output from Eq.~\ref{eq:suf_y} using the layer edge mask, we employ the exact mutual information optimization objective as used in \cite{gnnexplainer, wang2022reinforced} to train the flow mask. Further comparisons with GNNExplainer are described in Sec.~\ref{sec:mutual_information_training_comparisons}.

\subsection{Evaluation Metrics} \label{evaluation_metrics}

The main metrics we apply are Fidelity+, Fidelity-~\cite{survey, yuan2021explainability}, and Sparsity~\cite{survey}. According to the interpretation of \cite{amara2022graphframex}, we use Fidelity+ and Sparsity to evaluate the effectiveness of necessary explanations and evaluate sufficient explanations with Fidelity- and Sparsity. Furthermore, we use full recall sparsity (Sec.\ref{sec:target_retrieval_ability}) to evaluate explainer performances on flow-related tasks with human-design ground-truths, \ie, BA-Traffic. 
% It is critical to distinguish two different Accuracy metrics. Specifically, the Accuracy used in this work is used with explanation ground truths, while the Accuracy used in \cite{wang2022reinforced} evaluates the original task performance after applying the explanation masks. 
It is critical to note that the Accuracy used in \cite{wang2022reinforced} is equivalent to Fidelity-$^{acc}$~\cite{survey} that is similar to Fidelity-$^{prob}$. To keep consistency, the Fidelity+ and Fidelity- we adopt are the probability version, \ie, Fidelity+$^{prob}$ and Fidelity-$^{prob}$ in \cite{survey}. We provide detailed introductions and explanations of these metrics in the supplementary materials.

% \begin{table*}[t]
% 	\centering
% % 	% \vspace{-0.5cm}
% 	\caption{Results of ablation studies of our method on GCNs. FlowX$^\dagger$ denotes our FlowX without game theory initial assessments and FlowX$^*$ denotes our FlowX without learning refinement.}
% 	\begin{tabular}{lccccccc}
% 		\toprule
% 		& BA-Shapes &  BA-LRP & ClinTox &   Tox21 &    BBBP &    BACE & Graph-SST2 \\
% 		\midrule
%      $\mbox{FlowX}^\dagger$ &             0.40\textpm 0.03 &             0.17\textpm 0.10 & 0.34\textpm 0.08 &             0.17\textpm 0.05 &             0.39\textpm 0.17 &             0.32\textpm 0.06 &             0.22\textpm 0.09 \\
% FlowX* & 0.41\textpm 0.00 &    \textbf{0.52}\textpm 0.00 &             0.32\textpm 0.07 & 0.19\textpm 0.04 & 0.49\textpm 0.11 & 0.50\textpm 0.06 & 0.26\textpm 0.11 \\
% FlowX &    \textbf{0.42}\textpm 0.01 & 0.51\textpm 0.01 &    \textbf{0.38}\textpm 0.06 &    \textbf{0.22}\textpm 0.04 &    \textbf{0.57}\textpm 0.11 &    \textbf{0.51}\textpm 0.03 &    \textbf{0.32}\textpm 0.13 \\
% 		\bottomrule
% 	\end{tabular}
% 	\label{tab:ablation}
% % 	% \vspace{-0.3cm}
% \end{table*}

\newcommand{\reducehspace}{@{\hspace{0in}}}
\newcommand{\setwidth}{0.250\linewidth}

\begin{figure*}[t]
	\centering
        \resizebox{1\textwidth}{!}{
	$\begin{array}{c@{\hspace{0in}}c@{\hspace{0in}}c@{\hspace{0in}} c@{\hspace{0in}} c}
	     \includegraphics[width=\setwidth]{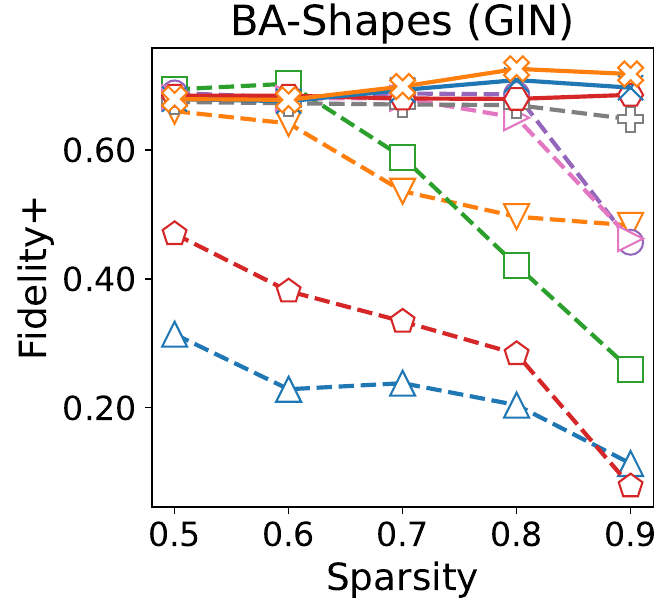}&
	     \includegraphics[width=\setwidth]{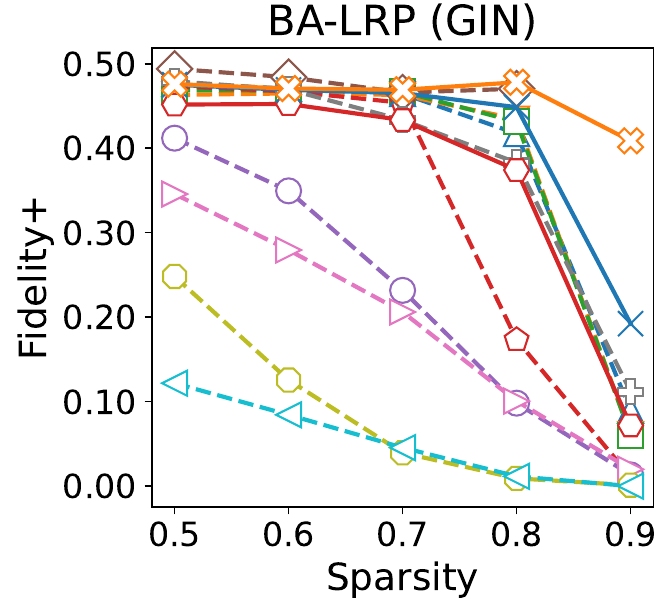}&
          \includegraphics[width=\setwidth]{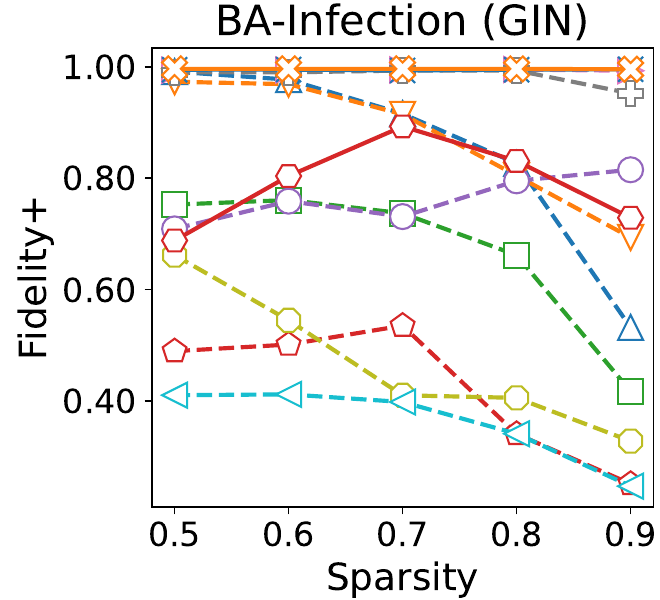}&
          \includegraphics[width=\setwidth]{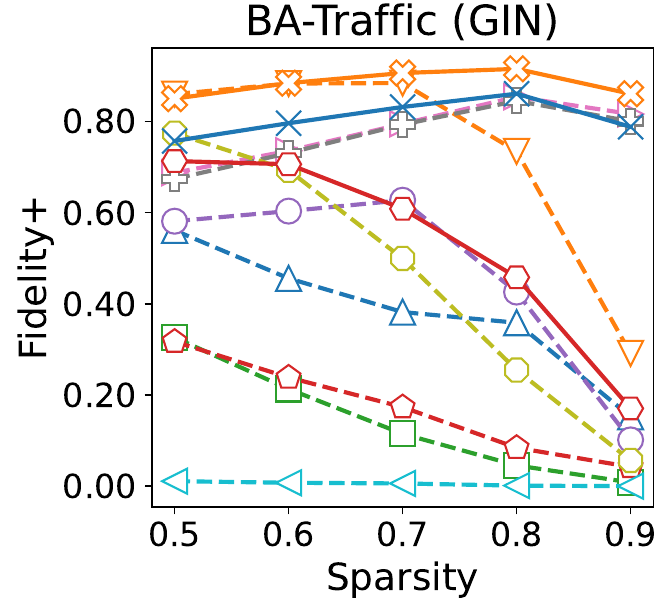}&
	     \includegraphics[width=\setwidth]{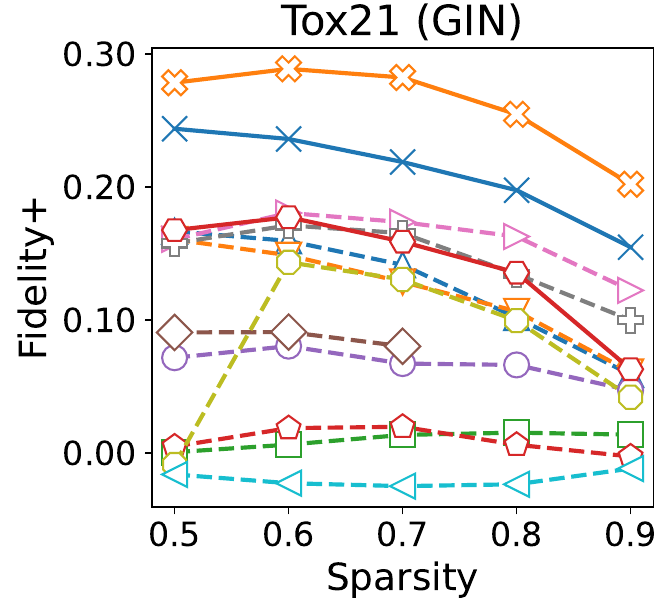} \\
	     \includegraphics[width=\setwidth]{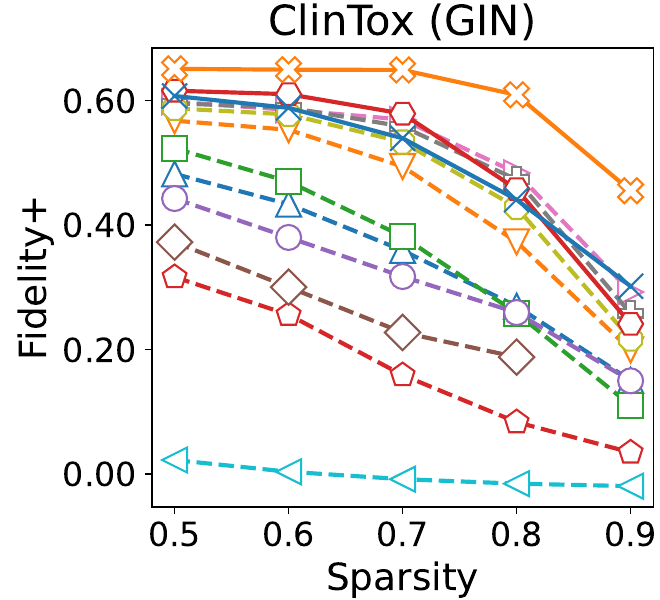}&
	     \includegraphics[width=\setwidth]{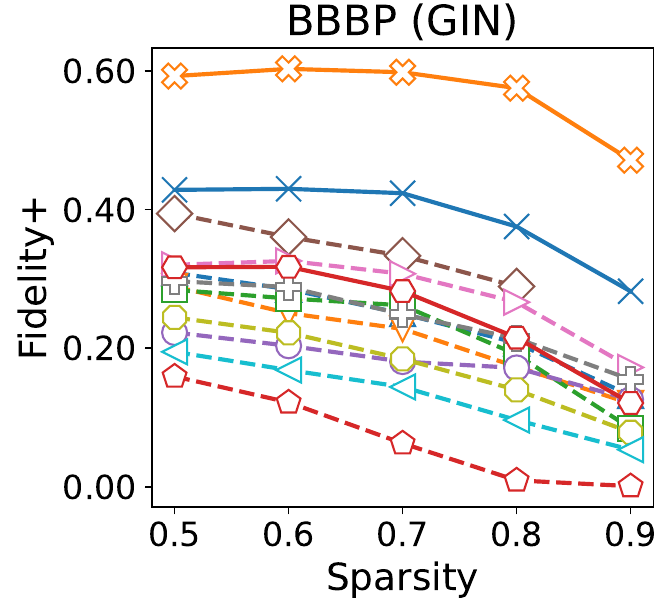}&
	     \includegraphics[width=\setwidth]{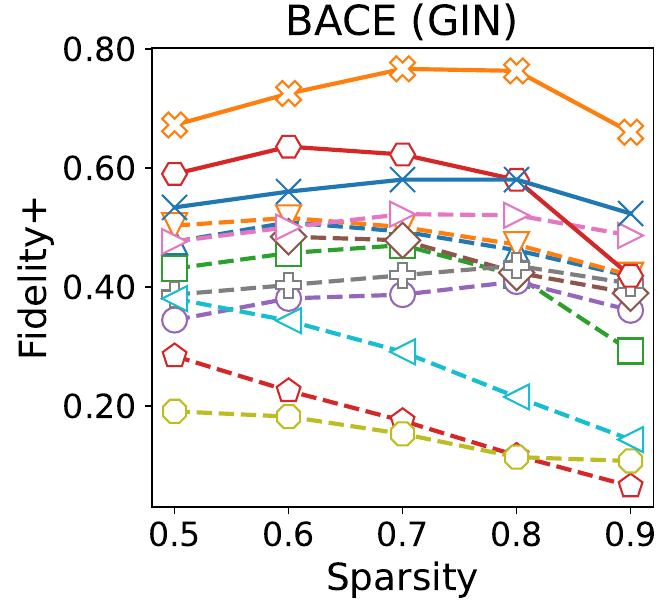}&
	     \includegraphics[width=\setwidth]{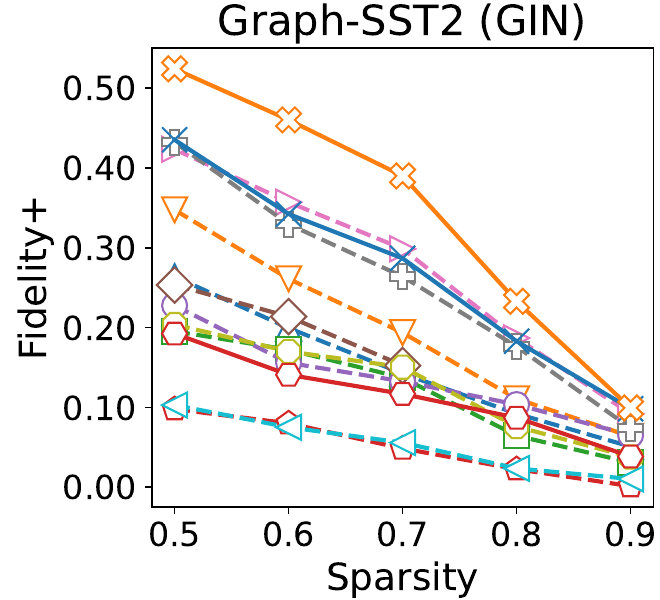}&
	     \includegraphics[width=\setwidth]{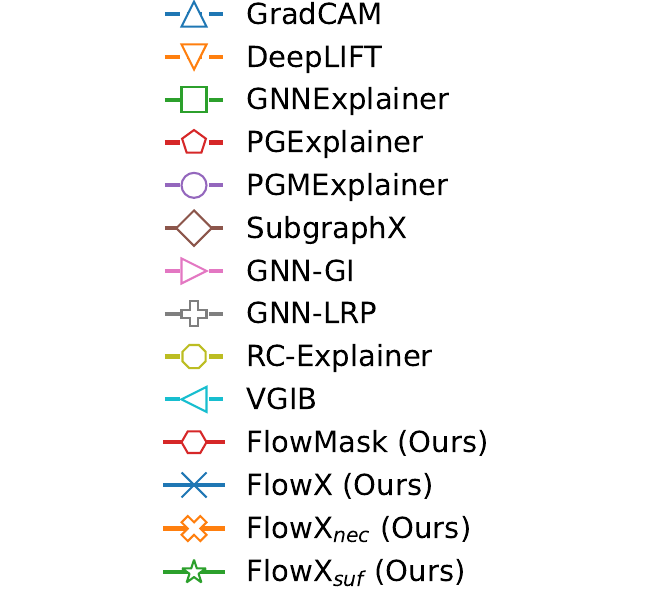}
	\end{array}$}
% 	\includegraphics[width=1\linewidth]{figures/exp1}% \vspace{-0.6cm}
    % % \vspace{-0.3cm}
	\caption{\textbf{Necessary explanation comparison.} We compare Fidelity+ values on 9 datasets with GINs under different Sparsity levels. Our methods are drawn in solid lines while baselines are drawn in dashed. Higher Fidelity+ indicates better performance.}
	\label{fig:fidelity+}
\end{figure*}

\begin{figure*}[t]
	\centering
        \resizebox{1\textwidth}{!}{
	$\begin{array}{c@{\hspace{0in}}c@{\hspace{0in}}c@{\hspace{0in}} c@{\hspace{0in}} c}
	     \includegraphics[width=\setwidth]{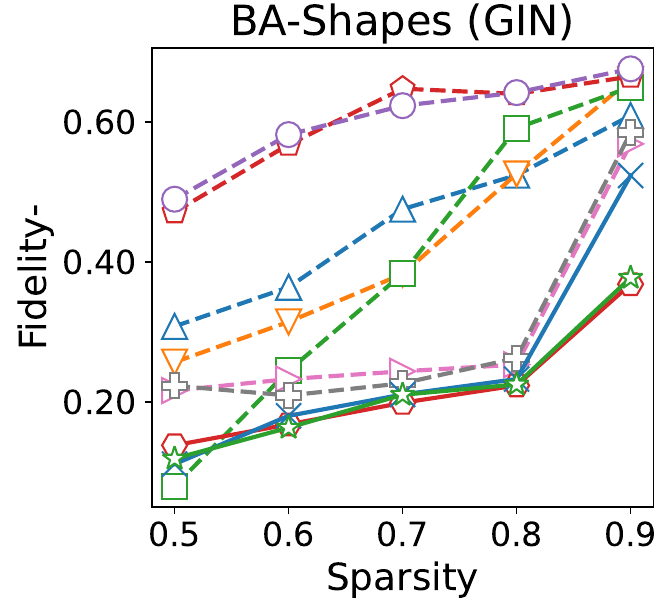}&
	     \includegraphics[width=\setwidth]{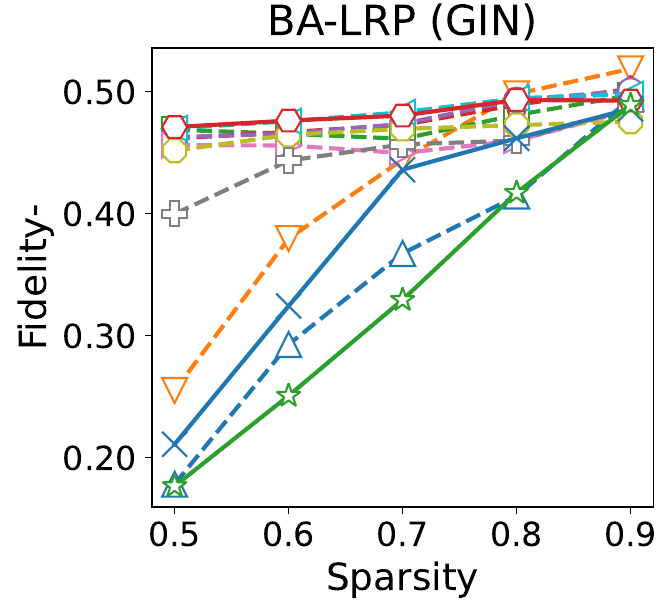}&
          \includegraphics[width=\setwidth]{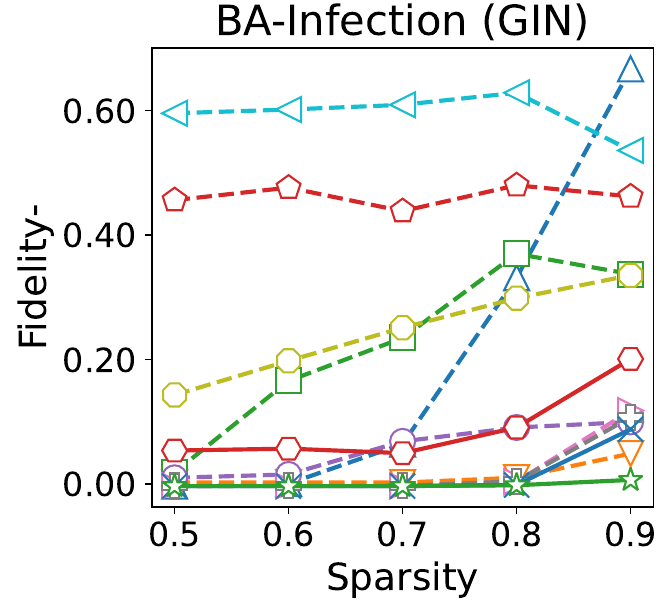}&
          \includegraphics[width=\setwidth]{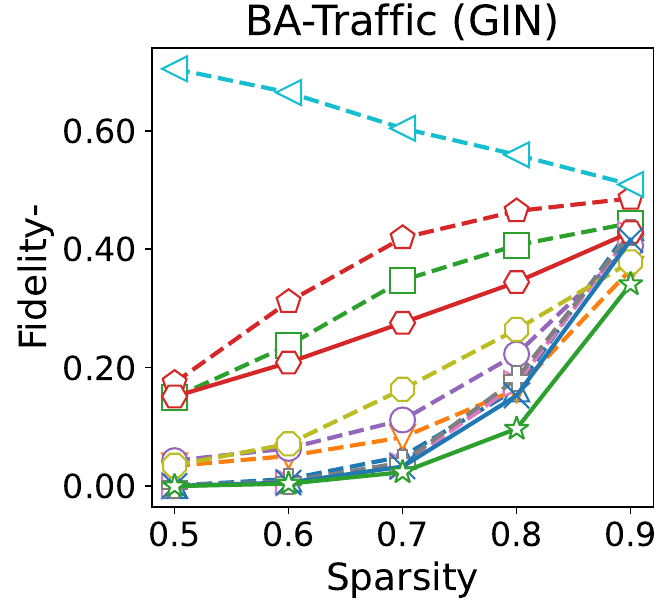}&
	     \includegraphics[width=\setwidth]{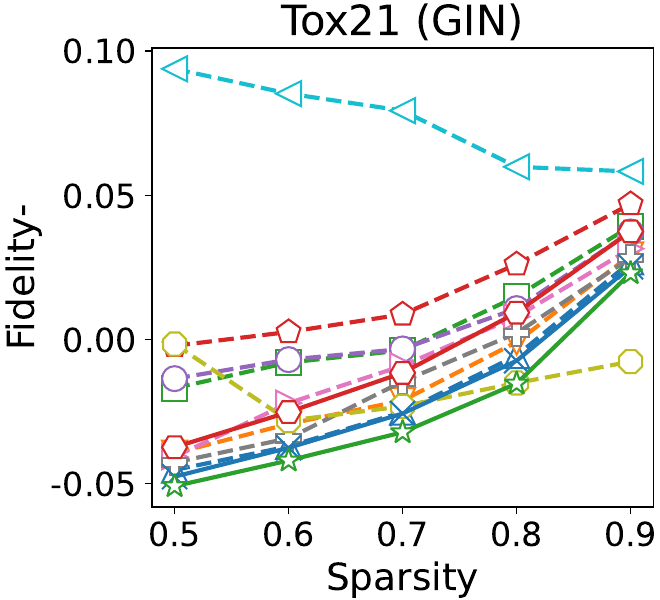} \\
	     \includegraphics[width=\setwidth]{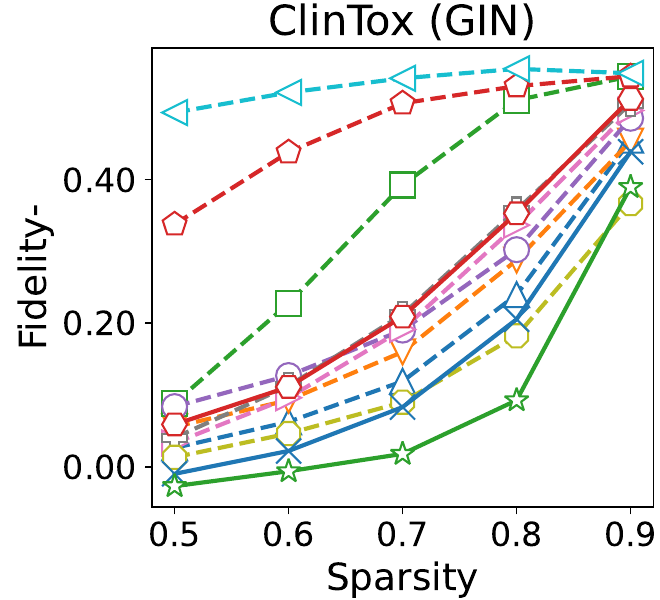}&
	     \includegraphics[width=\setwidth]{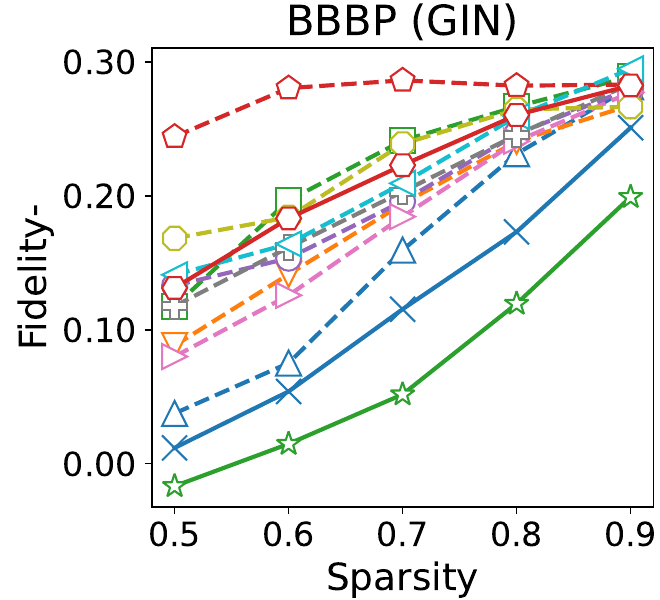}&
	     \includegraphics[width=\setwidth]{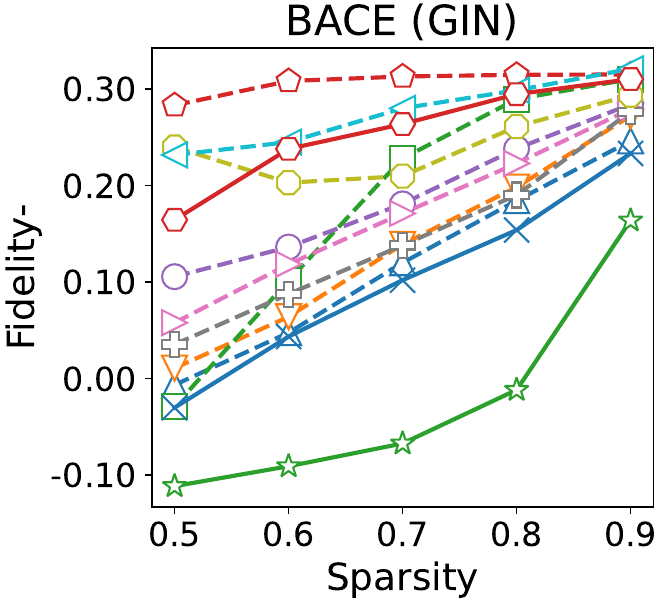}&
	     \includegraphics[width=\setwidth]{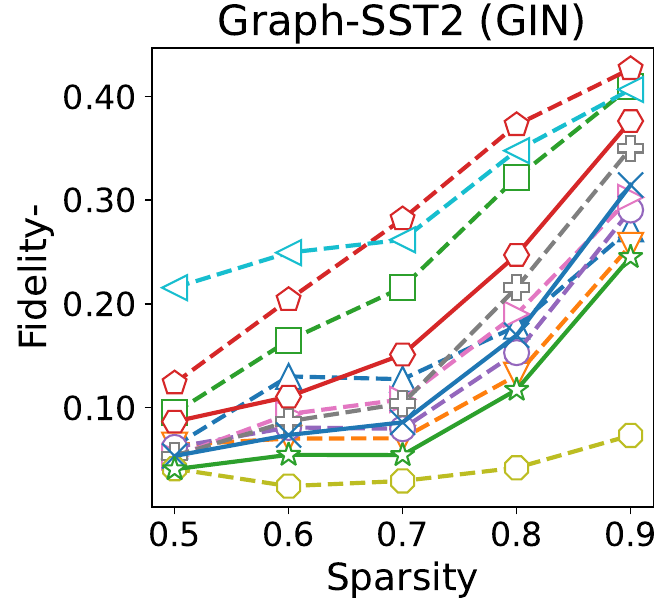}&
	     \includegraphics[width=\setwidth]{figures/results_plots/Fidelity-/legend1.pdf}
	\end{array}$}
% 	\includegraphics[width=1\linewidth]{figures/exp1}% \vspace{-0.6cm}
    % % \vspace{-0.3cm}
	\caption{\textbf{Sufficient explanation comparison.} We compare Fidelity- values on 9 datasets with GINs under different Sparsity levels. Our methods are drawn in solid lines while baselines are drawn in dashed. Lower Fidelity- indicates better performance.}
	\label{fig:fidelity-}
\end{figure*}

\subsection{Necessary Explanation Comparison}

We first quantitatively compare different explanation methods on Fidelity+ and Sparsity to answer the first research question. Good necessary explanations should be faithful to the model and capture comprehensive structures for the predictions. When such structures are removed, the original predictions should change significantly. In addition, a GNN model generally uses redundant information to make predictions.
% The Fidelity+ score measures the change of predicted probabilities when removing important input features identified by different explanation methods. 
Therefore, achieving high Fidelity+ scores indicates the removed structures are \textit{comprehensive} to the predictions. When comparing using Fidelity+, another desired property of explanations is Sparsity. To encourage the explanations to be more human-intelligible, they should contain fewer but more important features. Higher Sparsity scores indicate that fewer structures are identified as important in the explanations.
% Hence, we also employ the Sparsity metric which measures the percentage of input features that are identified as important. 
% It is noteworthy that these two metrics are highly correlated since the predictions tend to change more significantly when more input features are removed, \ie, explanations with higher Sparsity scores tend to have lower Fidelity+ scores. Hence,  we argue that the Fidelity+ scores need to be compared under a similar Sparsity level. 
Hence, we control the Sparsity scores of explanations and compare the corresponding Fidelity+ scores. For each dataset, we use samples from the test set and conduct such quantitative evaluations. The results are reported in Fig.~\ref{fig:fidelity+} where we show the plots of Fidelity+ scores with respect to different Sparsity levels. As shown in the figure, while FlowX (blue thin X) performs better than most methods, FlowX$_{nec}$ (orange hollow X) outperforms them significantly across all Sparsity levels. These stable performances on different datasets indicate the promising ability of our methods for various applications and generalization scenarios. For synthetic datasets, the comparisons are more clear under a high Sparsity level. 

% Meanwhile, our proposed method, GNN-LRP, and GNN-GI perform better than the other methods, which indicates the superiority of the  methods based on flows/walks. Note that GNN-LRP achieves good results on synthetic datasets but surprisingly, it is not better than GNN-GI on real-world datasets. In addition, PGExplainer only obtains competitive results on simple synthetic datasets but not on complex real-world datasets. Note that, for SubgraphX, its subgraph explanations are not directly comparable so we convert its subgraph explanations to edge explanations. 
% It performs well on the BA-LRP and BBBP datasets but not as expected on the other datasets. We believe the reason may be that such converting destroys the continuousness and completeness of its subgraph explanations.  
% Last but not least, we wish to mention that the comparisons using edge actually limit the performance of the flow-based methods due to explanation granularity changes.
% More results and time comparisons are also reported in the supplementary material~3.

\subsection{Sufficient Explanation Comparison}

In order to answer the second research question, we further conduct experiments on these datasets with metric Fidelity- and Sparsity for sufficient explanation comparisons. Intuitively, as the Sparsity increases, the selected subgraphs become smaller. If the explanation can preserve important structures to maintain the prediction performance, the explainer can achieve a low Fidelity-. As shown in Fig.~\ref{fig:fidelity-}, while FlowX (blue thin X) with pure flow samplings performs as a strong explainer, FlowX$_{suf}$'s performances (green star) surpass most methods on nearly all datasets clearly. The only exception happens on Graph-SST2 compared with RC-Explainer. There are two possible reasons behind this behavior. First, Graph-SST2, as a sentimental dataset, contains redundant sentimental information in sentences. This implies that only one or two keywords can be *sufficient* to make predictions, while large amounts of words need to be removed to change the predictions. Second, RC-Explainer reinforcement searches subgraphs from zero nodes, in which this tree-like searching process provides exhaustive-like searching for subgraphs with few nodes because of the smaller searching space near the roots of the searching tree. This exhaustive small subgraph searching behavior coincides with the need for sufficient explanations of Graph-SST2, leading to its high sufficient explanation performance. However, because of this characteristic, RC-Explainer produces suboptimal results on datasets requiring large explanations and long-term correlations in Fig.~\ref{fig:fidelity-} and is shown hard to search for necessary explanations that target covering comprehensive important information evaluated by Fidelity+.

Note that VGIB performs as a remarkable interpretable method, but its explanation ability is limited with fixed GNN models.

\begin{figure}
    \centering
    \resizebox{1\linewidth}{!}{
    $\begin{array}{c@{\hspace{0in}} c}
    \includegraphics{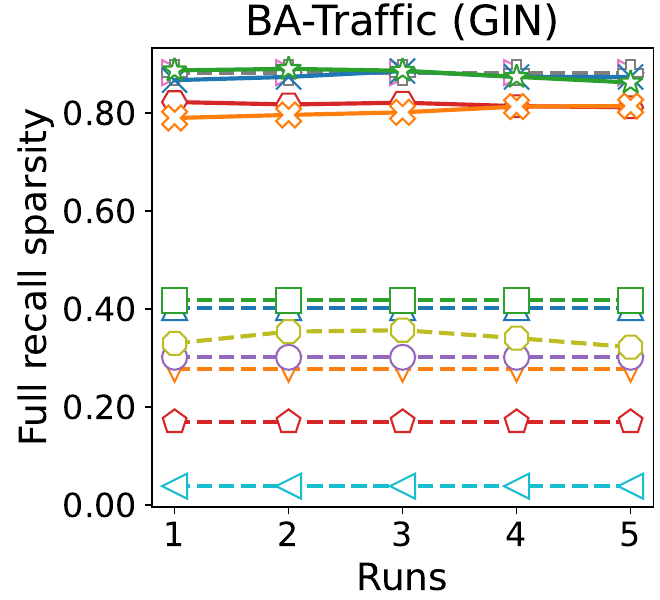} &
    \includegraphics{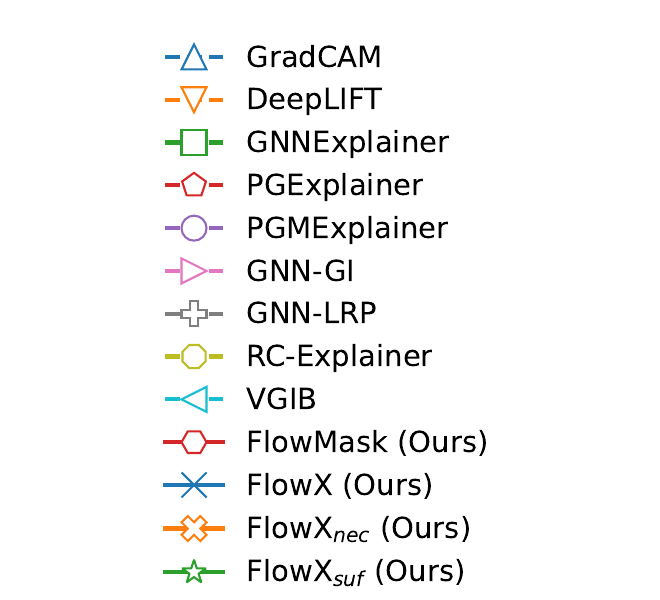}
    \end{array}$}
    \caption{Full recall sparsity comparison on BA-Traffic.}
    \label{fig:traffic}
\end{figure}

% \subsubsection{Human-intelligible explanation comparison on BA-Infection}

% BA-Infection, as described in the supplementary material, is a synthetic dataset that uses multi-hop "infection" correlations to determine labels. The ground truths of samples are the edges of the infection routes. We measure the accuracy of the predictions across different Sparsity as shown in Fig.~\ref{fig:infe_traffic}. The results provide two insights. First, flow-based methods, including GNN-GI and GNN-LRP, outperform all other methods. Second, FlowX, as a trade-off between sufficient and necessary explanations, produces the best results, especially with high Sparsity. It further implies that the model faithful explanations may not be human-intelligible explanations. Among the Shapley explanations, necessary explanations, and sufficient explanations, Shapley explanations are the most human-intelligible one, while the other two explanations are faithful to the model from different perspectives.

\subsection{Flow-based v.s. Non-flow-based Methods}

To demonstrate the benefits of using fine-grained flow-based methods and answer the third research question, we conduct experiments from the following two aspects.

\subsubsection{Target Edge/Flow Retrieval Ability}\label{sec:target_retrieval_ability}

BA-Traffic is a dataset simulating the challenge in Fig.~\ref{fig:challenge}. The base graphs of samples in BA-Traffic are 20-node 19-edge Barab$\acute{\textrm{a}}$si-Albert graphs that are constructed as traffic maps, where nodes are intersections, and edges are roads. In each sample, the node with the highest degree is a congested intersection. There are two types of traffic flows (car flows and bike flows), indicated by node features, going across the congested intersection and causing traffic jams. The label of the task is determined by the number of flows of each type, \ie, if the number of bike flows is larger than the number of car flows, the sample has label 0, otherwise, 1. There can be one or two flows for each type of flow, but the number of them cannot be the same. We provide the flow ground truths represented by edge sequences.

The explanation task is to cover all the true traffic flows, while we convert the true flows into edges for other explainers. Specifically, we calculate the lowest ratio between the number of selected edges/flows and the total number of edges/flows such that the selected edges/flows can cover all ground-truths. We denote ``1 - this ratio'' as a metric: full recall sparsity. Higher full recall sparsity indicates better retrieval performance. As shown in Fig.~\ref{fig:traffic}, flow-based methods have the ability to retrieve the real flows with high sparsity. However, other methods fail in this task from two aspects. First, they cannot retrieve true traffic flows intrinsically leaving unspecified edge explanations as depicted in Fig.~\ref{fig:challenge}. Second, the edge full recall sparsity is lower than the flow full recall sparsity, which indicates the flow explanations are more specified and concrete than the edge explanations. That is, edge explanations need larger portions of graphs to convey the same or commonly less information, compared to flow explanations. Therefore, flow explanations are more concrete and informative.

Note that the relatively weak performances produced by node-based explainers indicate the non-trivial information loss by converting node explanations to edge explanations.

\subsubsection{Mutual Information Training Comparisons}~\label{sec:mutual_information_training_comparisons}

Except for the benefits of the flow explanation itself, flow-based explainers have a unique advantage. This advantage originally comes from the flow modeling introduced in Sec.~\ref{sec:causal}, in which flow-based graph modeling can model multi-hop correlations while edge-based (graphon) modeling cannot. Empirically, to explore the merit brought from the modeling, we compare our flow-based training method FlowMask (Sec.~\ref{sec:flowmask}) with the similar edge-based method GNNExplainer in our comprehensive explanation experiments to show its favorable explanation performance.

FlowMask has the same mutual information training objective as GNNExplainer. The only distinction between them is the granularity of the trainable mask, \ie, a flow mask or an edge mask. As shown in Figs.~\ref{fig:fidelity+}, FlowMask (red hexagon, solid line) clearly outperforms GNNExplainer (green rectangular, dashed) on 7 out of 9 datasets for necessary explanation comparisons. For sufficient explanation comparisons, FlowMask's performances surpass GNNExplainer's non-trivially on 6 out of 9 datasets. Except for the Fidelity- comparison on BACE, FlowMask performs similarly to GNNExplainer on the rest of 4 datasets, \ie, BA-LRP and Graph-SST2 for Fidelity+; BA-LRP, BBBP for Fidelity-. Therefore, this experiment indicates that for the same training strategy, training on finer-grained explanation targets can produce better results, which implies the effectiveness of using flow-based methods from a training aspect.

According to the visualization in the supplementary materials and previous experimental results, we can obtain an insight: multi-hop correlation modeling and explanations can provide natural, more specified, and stronger explanation results.

% visualization: bbbp GCN 0.8 30 42, clintox 0.8 16; 0.7 24, 

\section{Conclusions}\label{conclusion}
We study the explainability of deep graph models, which are generally treated as black boxes. From the inherent functional mechanism of GNNs, we propose FlowX and its variants to explain GNNs by studying message flows. This work covers flow-based explanations from a systematical framework including an innovative flow-based graph modeling to comprehensively empirical implementations. Extensive experiments not only demonstrate the effectiveness and flexibility of FlowX and its variants but also emphasize the superiority of flow explanations. We hope this work can shed light on further multi-hop correlation modeling and explanation research.

\ifCLASSOPTIONcaptionsoff
  \newpage
\fi



\bibliographystyle{IEEEtran}
\bibliography{ref.bib}


% \end{thebibliography}

\begin{IEEEbiography}[{\includegraphics[width=1in,height=1.25in,clip,keepaspectratio]{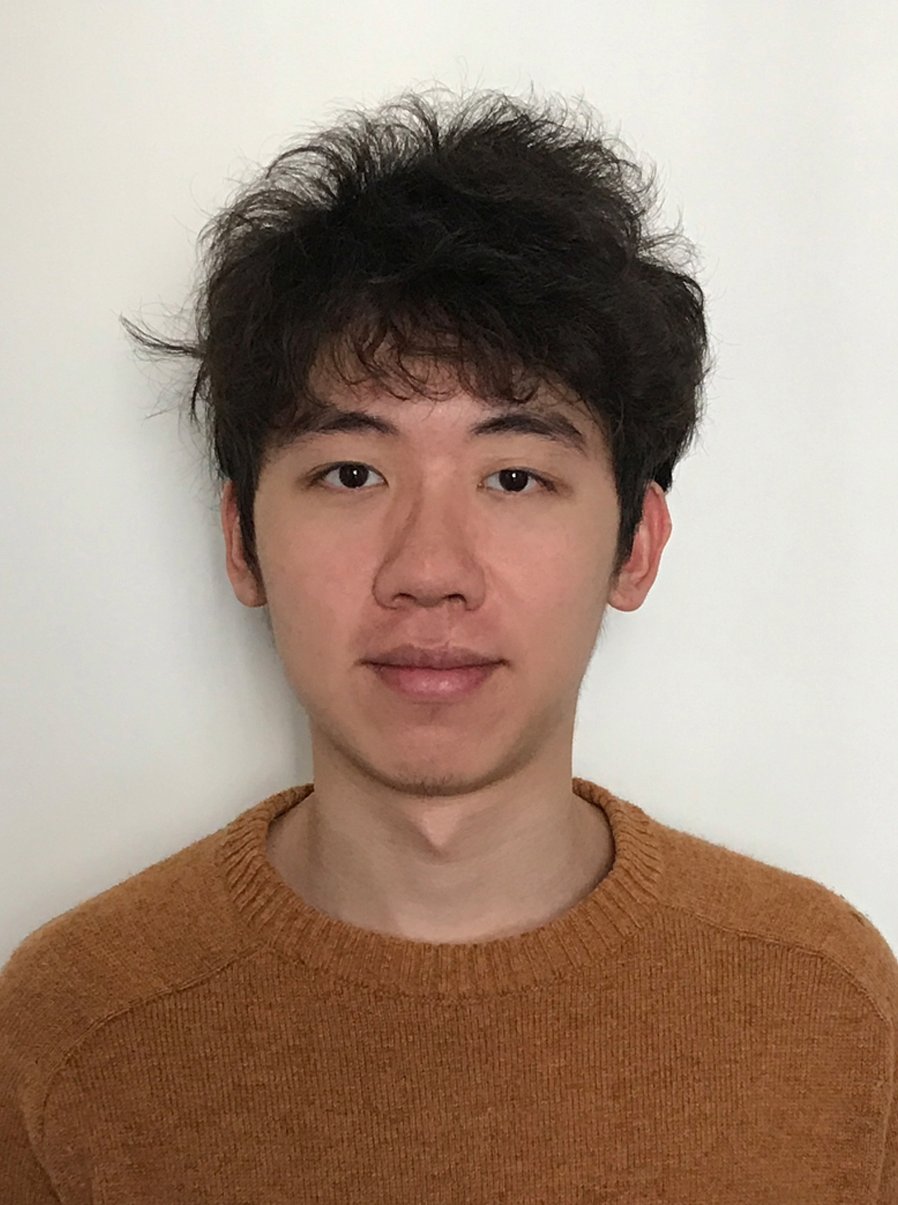}}]
    {Shurui Gui} received his B.S. in Computer Science from the University of Science and Technology of China in 2020. Currently, he is a Ph.D. student in Texas A\&M University, College Station, Texas. His research interests include deep learning, explainability, OOD generalization, and causality.
\end{IEEEbiography}

\vspace{-0.5cm}

\begin{IEEEbiography}[{\includegraphics[width=1in,height=1.25in,clip,keepaspectratio]{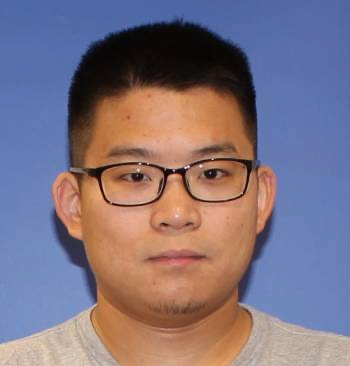}}]
    {Hao Yuan}is currently a research scientist at Meta Platforms, Inc. He received his PhD degree in Computer Science from Texas A\&M University, College Station, Texas, in 2021. Prior to this, he received his B.S. in Computer Science from the University of Science and Technology of China in 2012 and his M.S. in Computer Science from the Memorial University of Newfoundland in 2016. His research interests include machine learning, deep learning, and explainability.
\end{IEEEbiography}

\vspace{-0.5cm}

\begin{IEEEbiography}[{\includegraphics[width=1in,height=1.25in,clip,keepaspectratio]{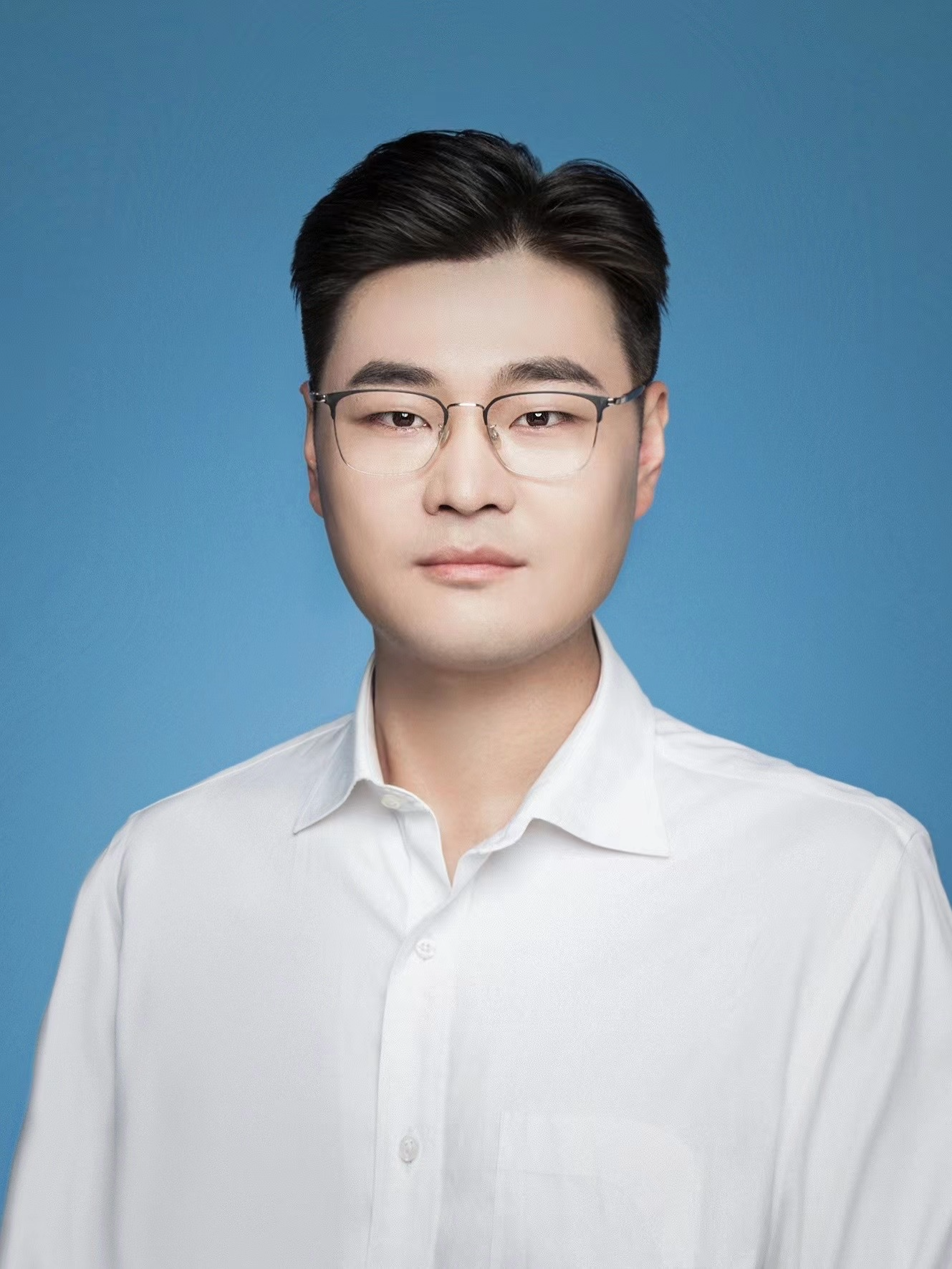}}]
{Jie Wang} received the B.Sc. degree in electronic information science and technology from University of Science and Technology of China, Hefei, China, in 2005, and the Ph.D. degree in computational science from the Florida State University, Tallahassee, FL, in 2011. He is currently a professor in the Department of Electronic Engineering and Information Science at University of Science and Technology of China, Hefei, China. His research interests include reinforcement learning, knowledge graph, large-scale optimization, deep learning, etc. He is a senior member of IEEE.
\end{IEEEbiography}

\vspace{-0.5cm}

\begin{IEEEbiography}[{\includegraphics[width=1in,height=1.25in,clip,keepaspectratio]{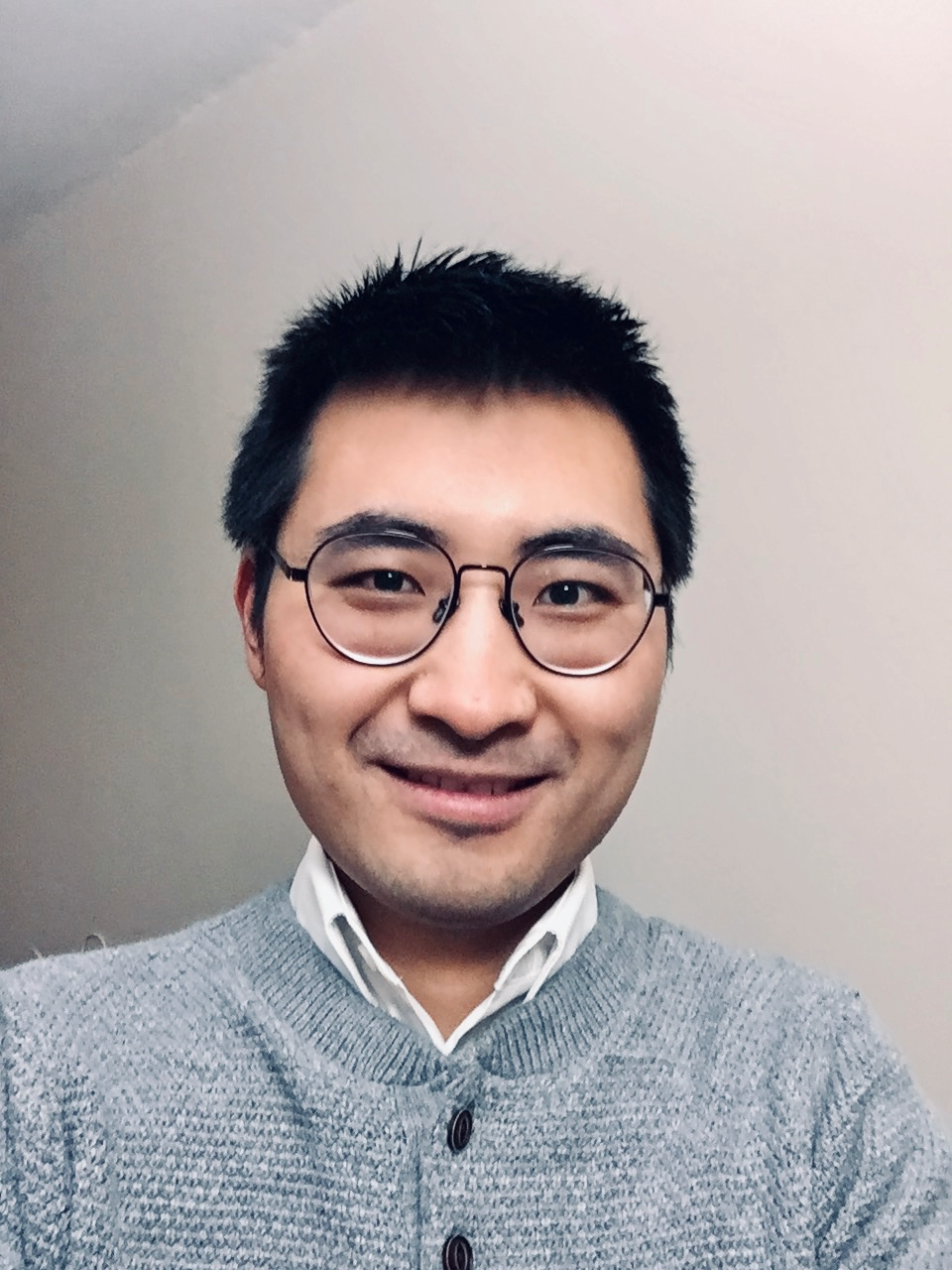}}]
{Qicheng Lao} received his B.S. degree in medicine from Fudan University, China, his M.Sc. degree in experimental medicine from McGill University, Canada, and his Ph.D. degree in computer science from Concordia University, Montreal. He is currently a post-doctoral fellow with the Montreal Institute for Learning Algorithms (MILA). His research interests include multimodal representation learning, and machine learning methods applied to healthcare.
\end{IEEEbiography}

\vspace{-0.5cm}

\begin{IEEEbiography}[{\includegraphics[width=1in,height=1.25in,clip,keepaspectratio]{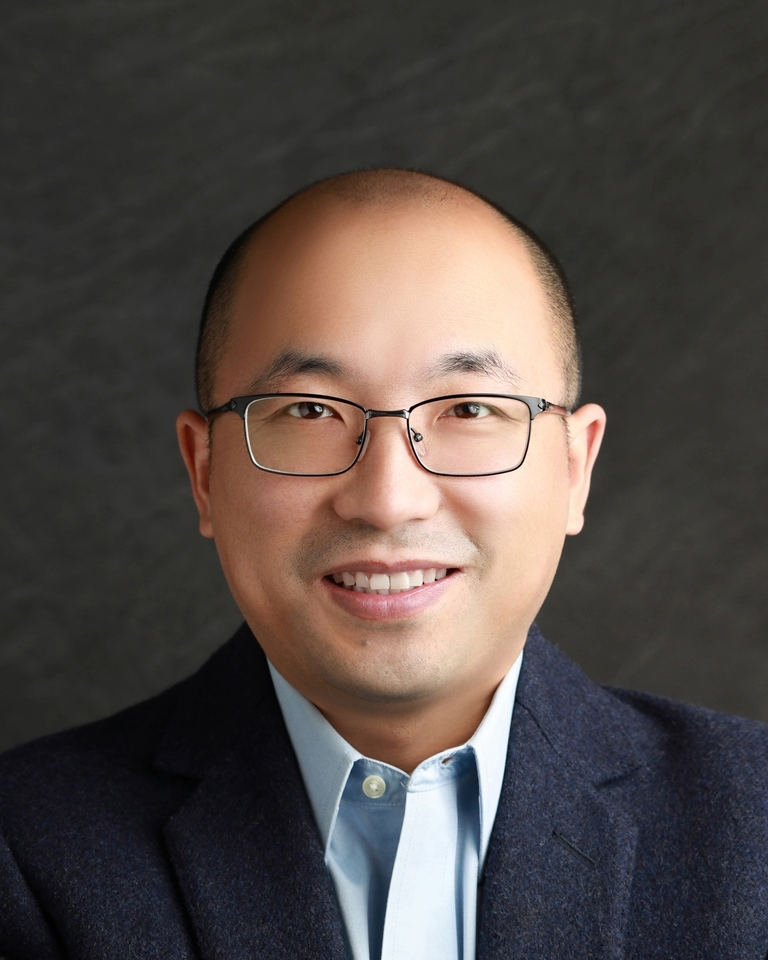}}]
{Kang Li} received the PhD degree in mechanical engineering from the University of Illinois at Urbana-Champaign, Champaign, IL, in 2009. He is an associate professor with the Department of Orthopaedics, New Jersey Medical School (NJMS), Rutgers University, Newark, NJ, and a graduate faculty member of the Department of Computer Science, Rutgers University.  His research interests include AI in healthcare, musculoskeletal biomechanics, medical imaging, healthcare engineering, design and biorobotics, and human factors/ergonomics.
\end{IEEEbiography}

\vspace{-0.5cm}

\begin{IEEEbiography}[{\includegraphics[width=1in,height=1.25in,clip,keepaspectratio]{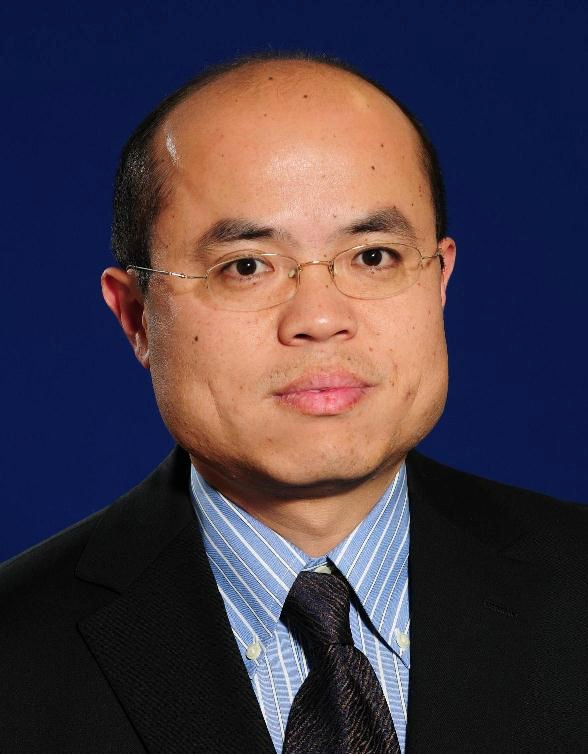}}]
{Shuiwang Ji} received the PhD degree in computer science from
Arizona State University, Tempe, Arizona, in 2010. Currently, he is
a Professor and Presidential Impact Fellow in the Department of
Computer Science and Engineering, Texas A\&M University, College
Station, Texas. His research interests include machine learning and
AI for science. He received the National Science Foundation CAREER
Award in 2014. He is currently an Associate Editor for IEEE
Transactions on Pattern Analysis and Machine Intelligence, ACM
Transactions on Knowledge Discovery from Data, and ACM Computing
Surveys. He regularly serves as an Area Chair or equivalent roles
for data mining and machine learning conferences, including AAAI,
ICLR, ICML, IJCAI, KDD, and NeurIPS. He is a Fellow of IEEE and
AIMBE, and a Distinguished Member of ACM.
\end{IEEEbiography}


% \end{IEEEbiography}


% % insert where needed to balance the two columns on the last page with
% %\newpage



%\vfill



\clearpage

\setcounter{section}{0}

\begin{strip}
\begin{center}
    \LARGE \bf {FlowX: Towards Explainable Graph Neural Networks via Message Flows Supplementary Materials}
\end{center}
\end{strip}

% \IEEEraisesectionheading{\section{Message Flow Intuition}\label{app:intuition}}
\section{Message Flow Intuition}\label{app:intuition}

\begin{figure*}[t!]
	\centering
	\includegraphics[width=1\linewidth]{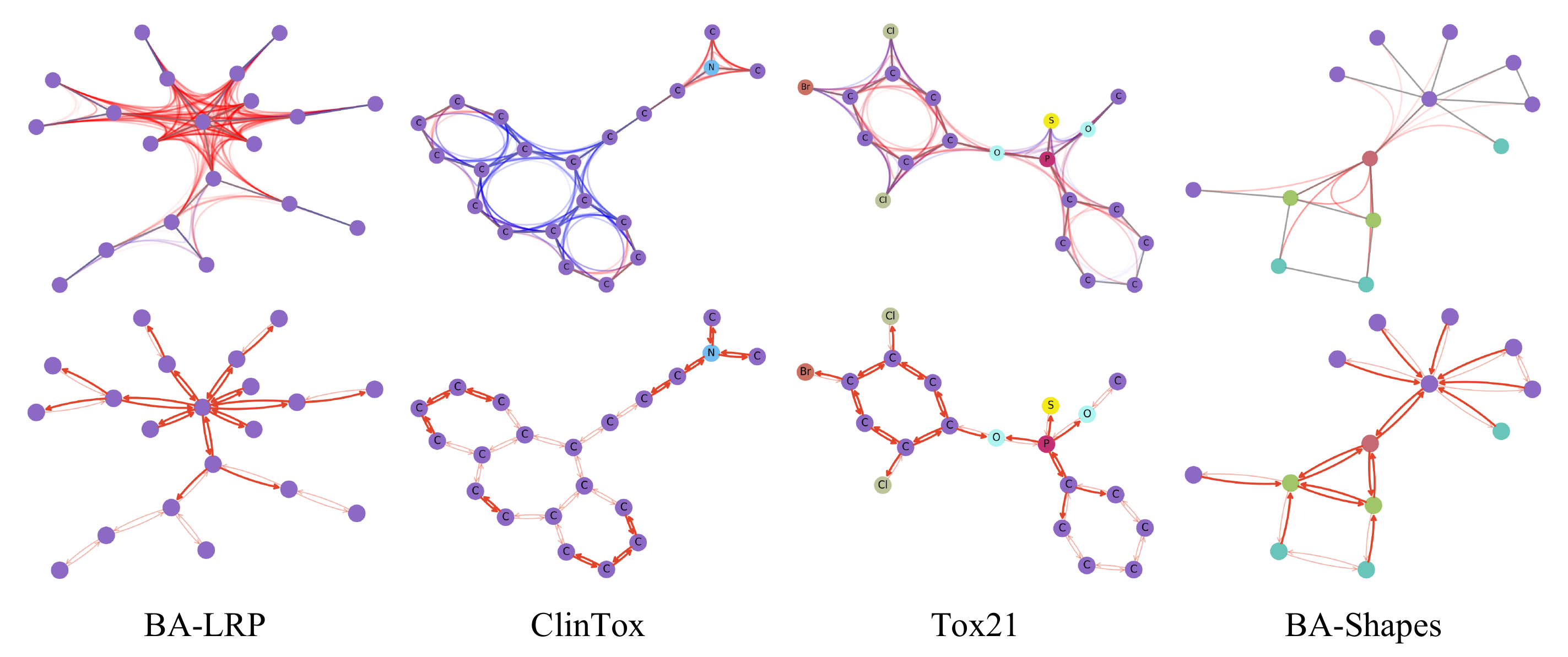}\vspace{-0.8cm}
	\caption{\small An illustration showing how to convert message flow contribution to edge contribution. The top row shows the 
	flow view of explanations while the bottom row shows the corresponding edge view. In the top row, we use red and blue flows to denote the positive and negative contributions respectively. In the bottom row, bold red arrow lines denote 
    important edges.}
	\label{fig:datasets}
\end{figure*}

\IEEEPARstart{M}{essage} flows can provide multi-hop dependence association. Take a virus infection dataset as an example. The reason why a given node is labeled as "susceptible" will be explained by a message flow beginning from an "affected" node to the given node. Therefore, we can provide an explanation that the message flow transmits the "affected" information from the "affected" node to the given node making the node a "susceptible" one. The upper figures of Figure.~\ref{fig:datasets} show how a node's message can be transmitted from the node to other nodes through different paths.

\section{Experimental Settings}\label{experimental_setting}

\subsection{Deep Graph Models}
We first introduce the details of the GNN models we try to explain. For all GNN models, we employ two message-passing layers for node classification (BA-Shapes~\cite{gnnexplainer}) and three for graph classification tasks (ClinTox, Tox21, BBBP, BACE~\cite{moleculenet}, BA-LRP~\cite{gnnlrp}, Graph-SST2~\cite{survey}). In addition, the final classifier consists of two fully connected layers.
For graph classification tasks, the average pooling is used to convert node embeddings to graph embeddings. In addition, the feature dimensions of message-passing layers and fully-connected layers are set to 300. We apply the ReLU function as the activation function after each message passing layer and fully-connected layer. In addition, the dropout is applied between two fully connected layers. Note that we consider both GCNs and GINs for all datasets. For the GCN layer, we employ the original normalized Laplacian matrix $\hat{A}^t=D^{-\frac{1}{2}}(A+I)D^{-\frac{1}{2}}$. For the GIN layer, two fully connected layers with the ReLU function are employed as the multilayer perceptron (MLP).

\subsection{Metrics}

Mathematically, a $K$-class classification dataset with $N$ samples can be represented as $\{((X_i, A_i), y_i)\ | \ i=1, 2, \ldots, N\}$, where $(X_i, A_i)$ is the input graph and $y_i$ is the ground-truth label of the $i$-th sample. The predicted class of a GNN model is $\hat{y}_i= \mathrm{argmax}f(X_i, A_i)$. Here the output of $f(X_i, A_i)$ is a $K$-length vector, in which each element denotes the probability of the corresponding class. Formally, we define the algorithm's explanation as a column vector $\dot{\mathbf{m}}_i$ that represents a mask on explainable targets where the number of targets is $|\dot{\mathbf{m}}_i|$. In this mask, the explanation method marks each selected target as a value $1$, otherwise a $0$. The source of the mask is a vector $\dot{\mathbf{c}}_i$ with the same shape that stores the contribution score of each explainable target. 
Intuitively, we prefer to assign targets with $1$s in the $\dot{\mathbf{m}}_i$ if the corresponding scores in $\dot{\mathbf{c}}_i$ are relatively high.
With these notations, we introduce the metrics employed for quantitative comparison.

\subsubsection{Fidelity+}

Following the metrics mentioned in the graph explainability survey~\cite{survey}, we choose the metric Fidelity+ to measure the faithfulness of the explainability methods to the model.
Mathematically, Fidelity+ is defined as:
\begin{flalign}
	\mbox{Fidelity+} &= \frac{1}{N}\sum_{i=1}^N ( f(X_i, A_i)_{\hat{y}_i} - f^{\overline{\dot{\mathbf{m}}}_i}(X_i, A_i)_{\hat{y}_i} ), \label{eq:fidelity+}
\end{flalign}
where $f^{\dot{\mathbf{m}}_i}(X_i, A_i)_{\hat{y}_i}$ denotes the output class $\hat{y}_i$'s probability when masking out the edges based on ${\dot{\mathbf{m}}_i}$. 
Specifically, we keep the value unchanged for the explainable targets that are marked as $1$ in $\dot{\mathbf{m}}$, while setting the other explainable targets as 0. Note that $\overline{\dot{\mathbf{m}}}_i = \mathbb{1} - \dot{\mathbf{m}}$ is complementary mask where $\mathbb{1}$ is an all-one vector that has the same shape as $\dot{\mathbf{m}}$. In the evaluation period, the higher Fidelity+ indicates the better performance of the explainability algorithm. Intuitively, Fidelity+ represents the predicted probability dropping after occluding the important features. Therefore, Fidelity+ can measure whether the chosen features are necessary to the predictions from the model's perspective.

\subsubsection{Fidelity-}

To evaluate whether the model can maintain the performances while only keeping the $\dot{\mathbf{m}}$ part, we can employ Fidelity- as follows:

\begin{flalign}
	\mbox{Fidelity-} &= \frac{1}{N}\sum_{i=1}^N ( f(X_i, A_i)_{\hat{y}_i} - f^{\dot{\mathbf{m}}}_i(X_i, A_i)_{\hat{y}_i} ). \label{eq:fidelity-}
\end{flalign}

A lower Fidelity- indicates a better sufficient explanation performance.

\subsubsection{Sparsity}\label{sec:sparsity}

In order to fairly compare different techniques, we argue that controlling the Sparsity~\cite{pope2019explainability, survey} of explanations is necessary.
Mathematically, Sparsity is defined as:
\begin{equation}
    \mbox{Sparsity}=\frac{1}{N}\sum_{i=1}^N \left(1-\dfrac{\sum_{k=1}^{|\dot{\mathbf{m}}_i|} \dot{m}_{ik}}{|\dot{\mathbf{m}}_i|}\right).
\end{equation}
A higher Sparsity score means fewer explainable targets are selected. Note that  the selections of targets are 
determined by the ranking of the contribution score in $\dot{\mathbf{c}}_i$.
It is noteworthy that Sparsity and Fidelity+/Fidelity- are highly correlated since explanations with higher Sparsity scores tend to
have lower Fidelity+ and higher Fidelity- scores.
For fair comparisons, we compare different methods with similar sparsity levels. 

\subsection{Training Settings}\label{app:training}
The models on different datasets use different learning rates and decay settings. Generally, we set the learning rate to $1\times 10^{-3}$ and the learning rate decay is equal to $0.5$ after 500 epochs. For datasets BA-LRP and Graph-SST2, the total number of epochs is 100 while we train models for 1000 epochs on the other datasets. All datasets are split into the training set (80\%), validation set (10\%), and testing set (10\%). All experiments are conducted using one NVIDIA 2080Ti GPU.

% All of the datasets' properties and their corresponding model training results on test sets are included in Table~\ref{tab:datasets}.
\begin{table*}[t]
	\centering
% 	\small
	\caption{\small The averaged time cost of eight algorithms. N/A denotes not applicable because PGExplainer, different from other explainers, begins explaining after training the whole dataset instead of explaining graphs one by one. Thanks to the training process, our time cost is controllable by decreasing the MC sampling times and relying more on the training process.} 
% 	\footnotesize
	\setlength{\tabcolsep}{0.9mm}{
		\begin{tabular}{lcccccccc}
			\toprule
			& GradCAM &  DeepLIFT & GNNExplainer &   PGExplainer &    GNN-GI &    GNN-LRP & SubgraphX &FlowX \\
			\midrule
			Time (ms) & 14 & 15 & 353 & N/A & 704 & 5993 & 407784 & 4501 \\
			\bottomrule
	\end{tabular}}
	\label{tab:time}
\end{table*}

% \begin{table}[t]
% 	\centering
% 	\caption{\small Comparisons between FlowX and three methods in terms of average Accuracy with 0.9 Sparsity on GCNs. We denote Accuracy, GNNExplainer, PGExplainer, PGMExplaienr as Acc., GNNEx., PGEx., PGMEx. respectively.}
% 	\begin{tabular}{lcccc}
% 		\toprule
% 		Acc. & GNNEx. &  PGEx. & PGMEx. &   FlowX  \\
% 		\midrule
% 		GCN &    0.2528 &  0.2199 &  0.3612 &  0.5266  \\
% 		GIN &    0.2917 &  0.2143 &  0.3748 &  0.4265 \\
% 		\bottomrule
% 	\end{tabular}
% 	\label{tab:accuracy}
% \end{table}

% \newcommand{\reducehspace}{@{\hspace{0in}}}
% \newcommand{\setwidth}{0.250\linewidth}

\begin{figure*}[t]
	\centering
        \resizebox{1\textwidth}{!}{
	$\begin{array}{c@{\hspace{0in}}c@{\hspace{0in}}c@{\hspace{0in}} c@{\hspace{0in}} c}
	     \includegraphics[width=\setwidth]{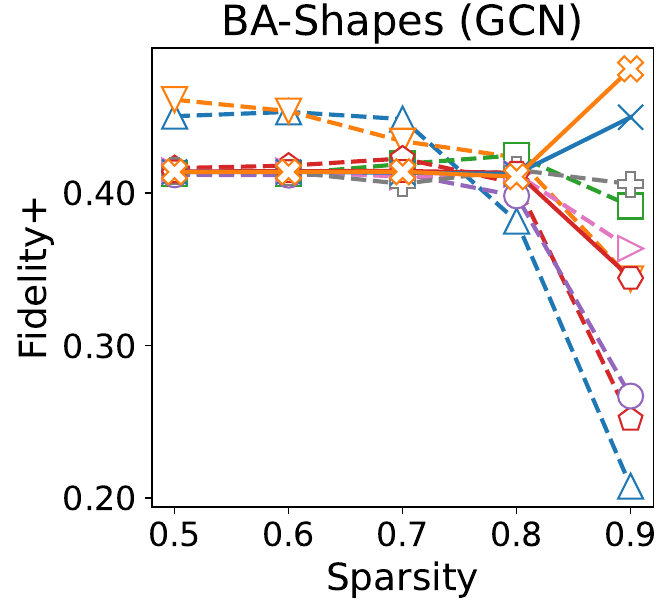}&
	     \includegraphics[width=\setwidth]{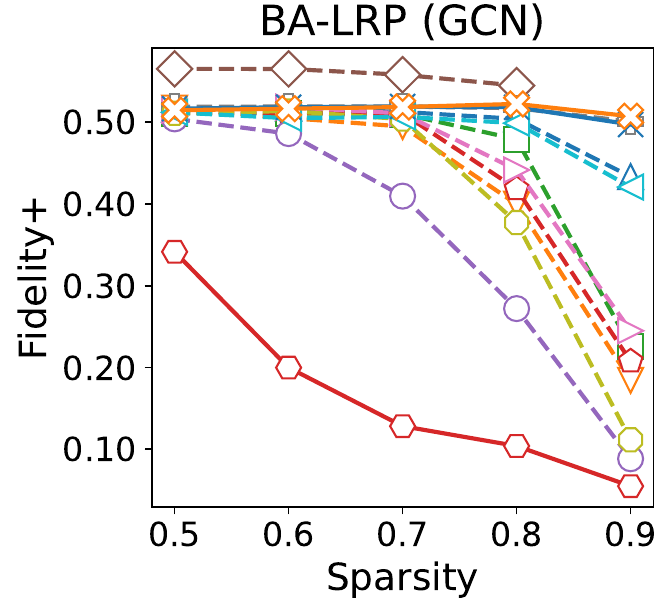}&
          \includegraphics[width=\setwidth]{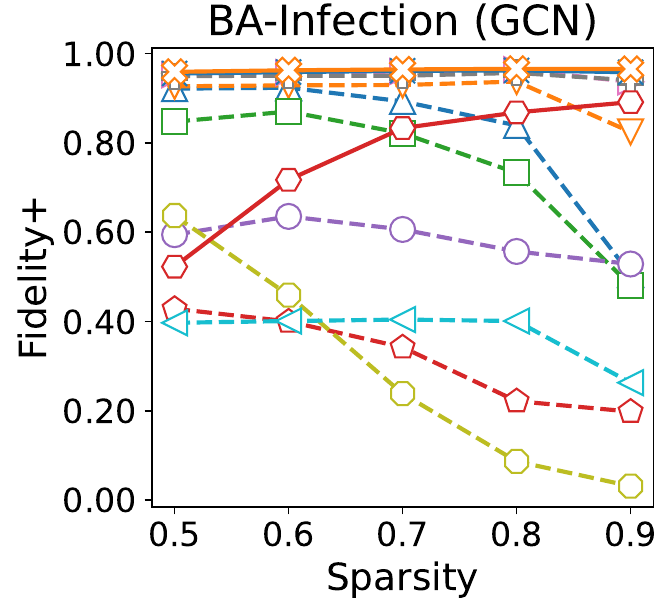}&
          \includegraphics[width=\setwidth]{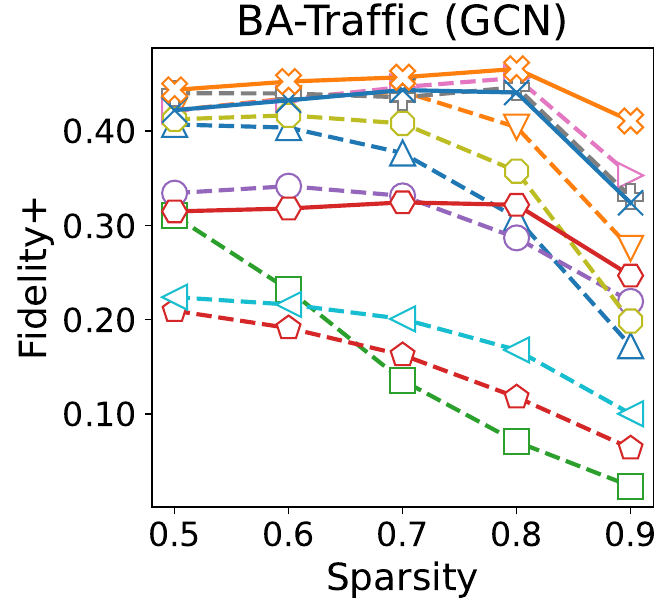}&
	     \includegraphics[width=\setwidth]{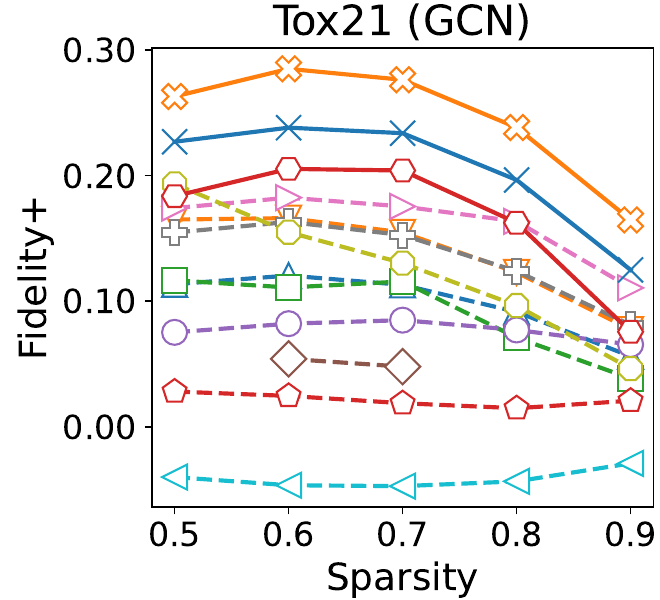} \\
	     \includegraphics[width=\setwidth]{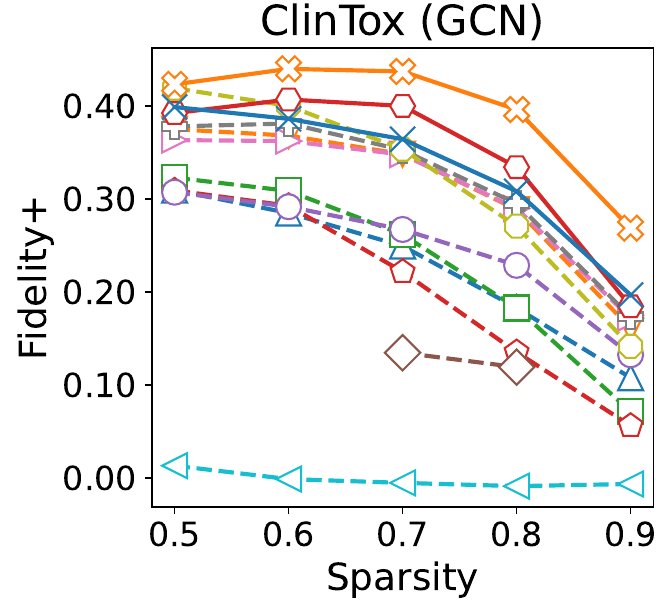}&
	     \includegraphics[width=\setwidth]{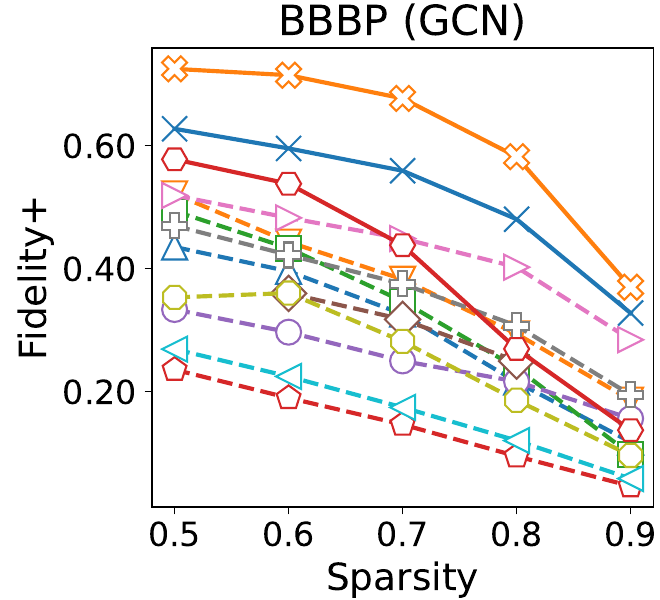}&
	     \includegraphics[width=\setwidth]{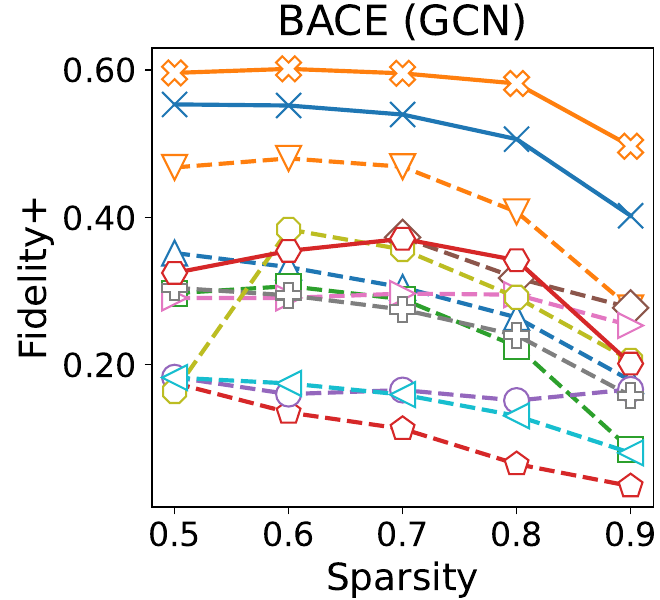}&
	     \includegraphics[width=\setwidth]{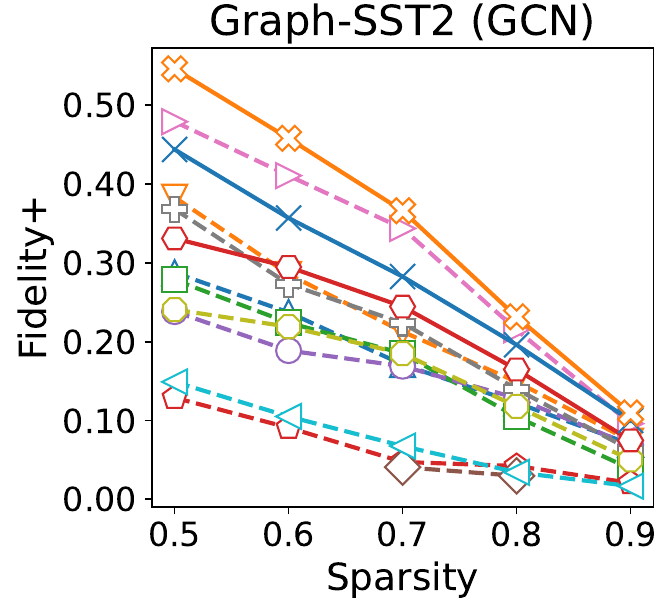}&
	     \includegraphics[width=\setwidth]{figures/results_plots/Fidelity-/legend1.pdf}
	\end{array}$}
% 	\includegraphics[width=1\linewidth]{figures/exp1}% \vspace{-0.6cm}
    % % \vspace{-0.3cm}
	\caption{\textbf{Necessary explanation comparison.} We compare Fidelity+ values on 9 datasets with GCNs under different Sparsity levels. Our methods are drawn in solid lines while baselines are drawn in dashed. Higher Fidelity+ indicates better performance.}
	\label{fig:gcn_fidelity+}
\end{figure*}

\begin{figure*}[t]
	\centering
        \resizebox{1\textwidth}{!}{
	$\begin{array}{c@{\hspace{0in}}c@{\hspace{0in}}c@{\hspace{0in}} c@{\hspace{0in}} c}
	     \includegraphics[width=\setwidth]{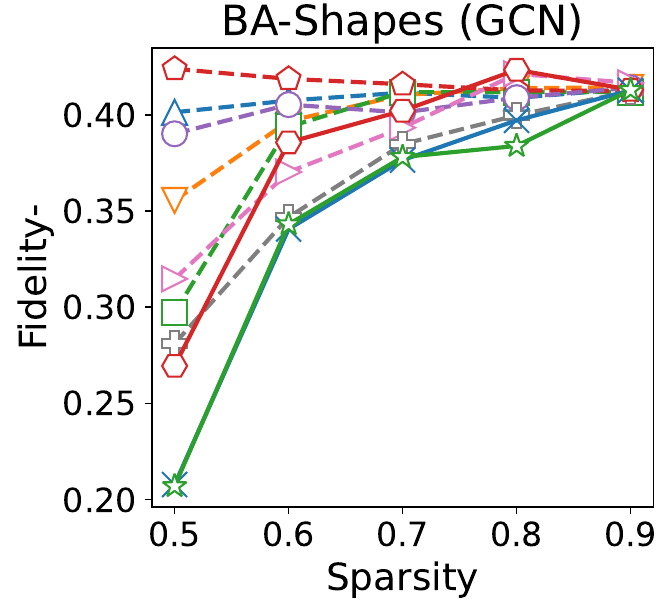}&
	     \includegraphics[width=\setwidth]{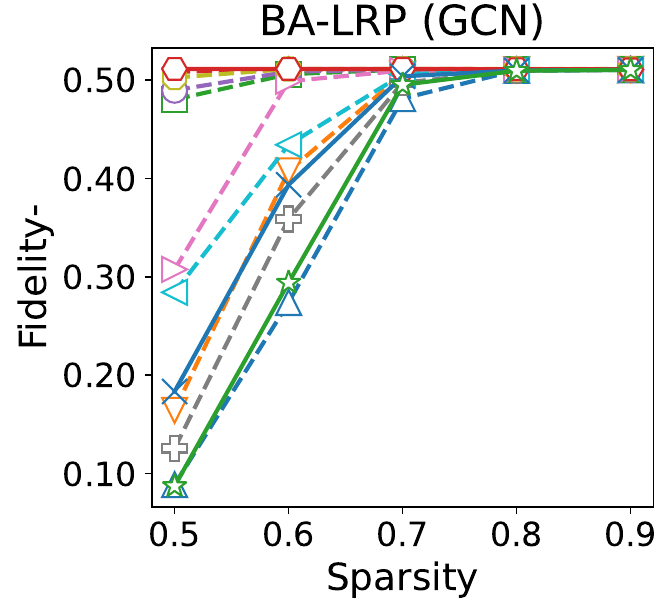}&
          \includegraphics[width=\setwidth]{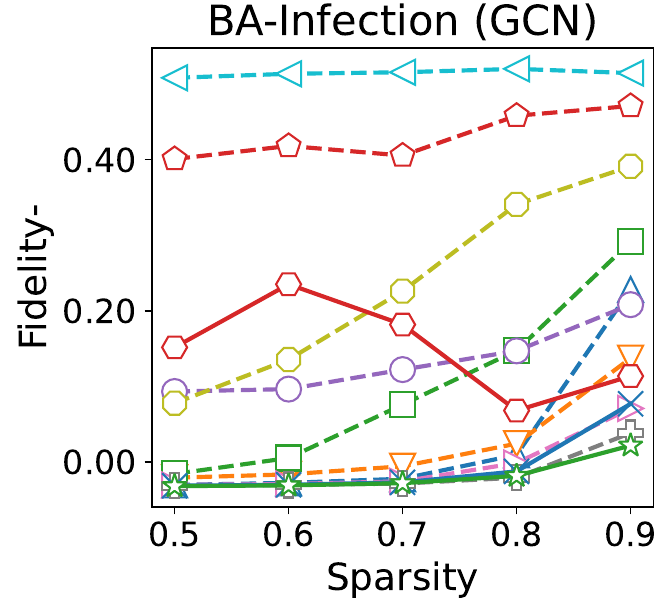}&
          \includegraphics[width=\setwidth]{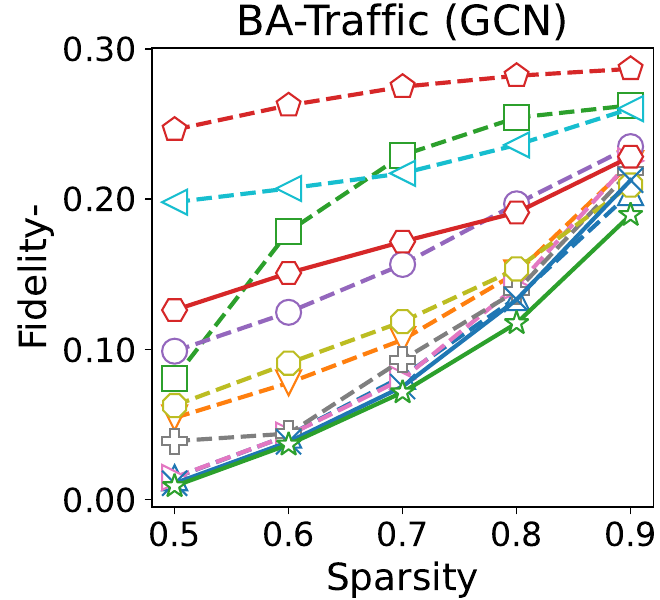}&
	     \includegraphics[width=\setwidth]{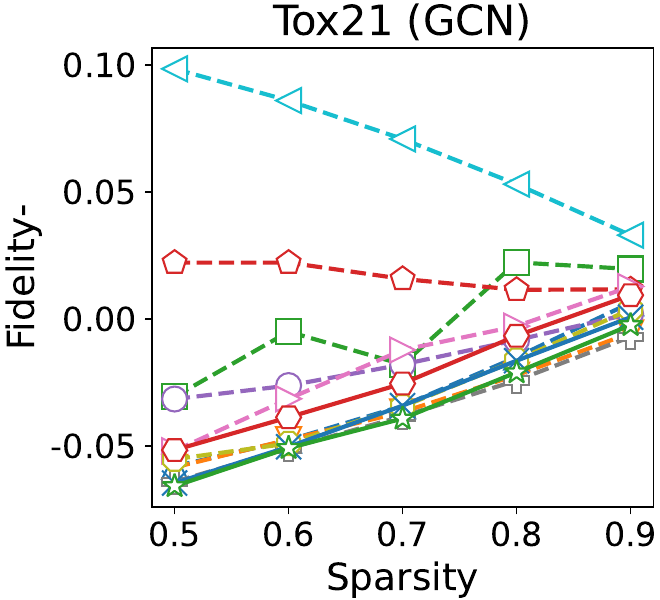} \\
	     \includegraphics[width=\setwidth]{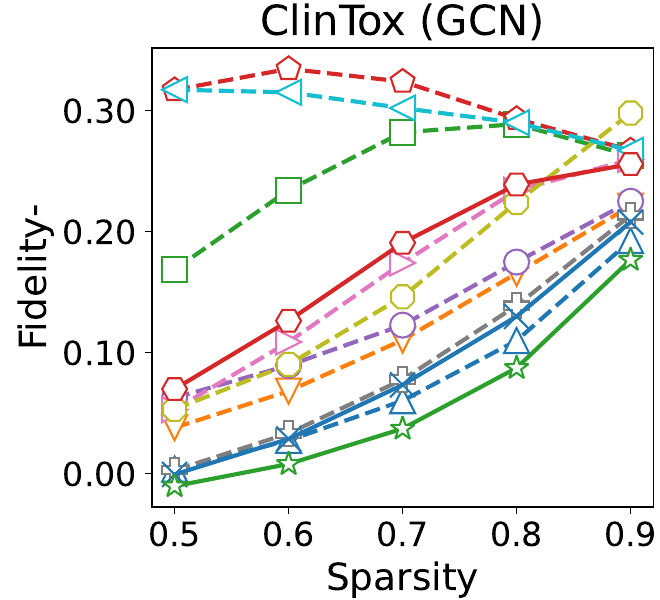}&
	     \includegraphics[width=\setwidth]{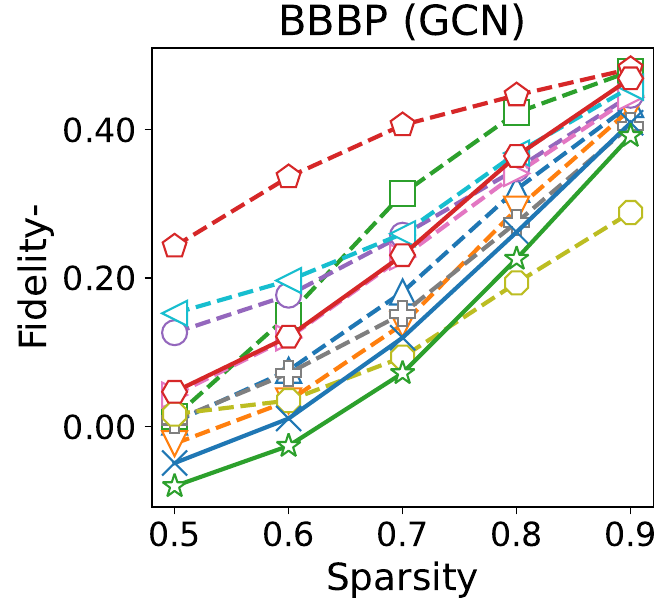}&
	     \includegraphics[width=\setwidth]{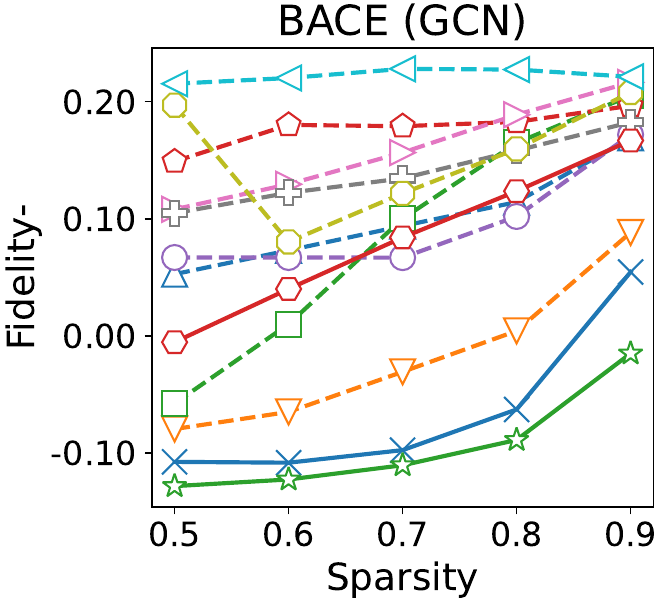}&
	     \includegraphics[width=\setwidth]{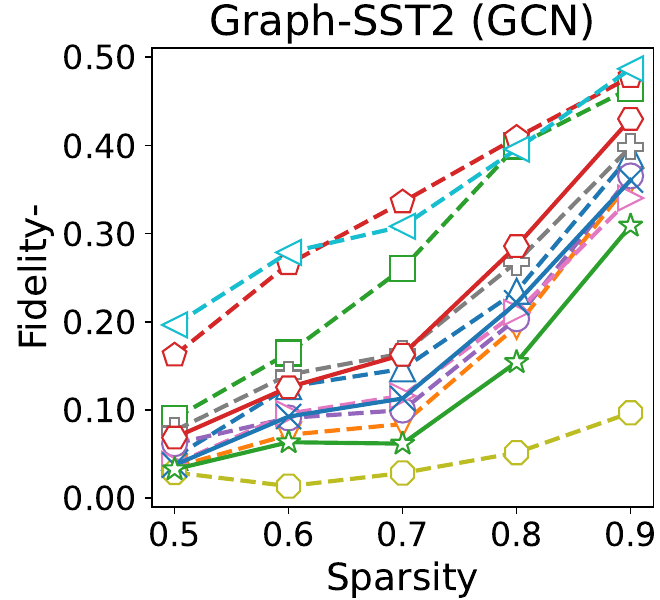}&
	     \includegraphics[width=\setwidth]{figures/results_plots/Fidelity-/legend1.pdf}
	\end{array}$}
% 	\includegraphics[width=1\linewidth]{figures/exp1}% \vspace{-0.6cm}
    % % \vspace{-0.3cm}
	\caption{\textbf{Sufficient explanation comparison.} We compare Fidelity- values on 9 datasets with GCNs under different Sparsity levels. Our methods are drawn in solid lines while baselines are drawn in dashed. Lower Fidelity- indicates better performance.}
	\label{fig:gcn_fidelity-}
\end{figure*}

\subsection{Targets of Explanations}

Since different techniques focus on different explainable targets, such as nodes, edges, flows, etc., these methods cannot be directly compared. For fair comparisons, we convert all explainable targets to graph edges. First, GNNExplainer and PGExplainer provide edge-level explanations so their results are directly used.  Second, for flow-based methods, including GNN-GI, GNN-LRP, and our FlowX, we convert the flow importance to edge importance by summing the total contribution of message flows that go through a particular edge. In addition, for node-based methods such as GradCAM~\cite{pope2019explainability} and DeepLIFT~\cite{deeplift},
the node importance is mapped to edge importance that the contribution of an edge is the averaged contribution of its connected nodes. Especially, for the subgraph-based method SubgraphX~\cite{yuan2021explainability}, we pick the explainable subgraph out, then assign the edges in this subgraph instead of nodes as the explanation. Note that the absolute values of contribution are not important since the metrics focus on their relative rankings. 
The Figure.~\ref{fig:datasets} shows how to map from message flows to edges.

\subsection{FlowX Settings}\label{app:flowx_setting}
At the first stage, the flow sampling time $M$ is set as $50$. In the second stage, we randomly initialize the weight vector of our learning refinement from the uniform distribution $[0, 0.1]$. During the training, we apply a learning rate of $10^{-1}$ without learning rate decay and train the weight vector for $500$ iterations.

\section{More Experimental Results}\label{app:experiment}

In this section, we provide additional quantitative comparison results. 

\subsection{Fidelity+ and Fidelity- Results on GCNs}

We report the plots of Fidelity+/Fidelity- scores with respect to different Sparsity levels for all datasets and GCN models in Figures~\ref{fig:gcn_fidelity+} and \ref{fig:gcn_fidelity-}. Clearly, the performance of our method is stable and better especially on the complex real-world datasets. 

\subsection{BA-Infection}\label{app:acc}

% To further demonstrate the effectiveness of our proposed method, we conduct experiments to compare the explanatory accuracy of three methods that usually evaluated in terms of accuracy. The results are reported in Table~\ref{tab:accuracy}. Since the important components are only a small portion of graphs, only high (0.9) Sparsity can compare these explainers clearly. These results are obtained from a synthetic graph classification dataset BA-INFE. 
BA-Infection is a synthetic graph classification dataset. Each graph includes a base Barab$\acute{\textrm{a}}$si-Albert graph and four types of motifs. Two classes of motifs are attached (2 motifs denote one property, 2 for another; let’s denote them as class/property A and B). We first connect the same number (1~3) of each class's motifs to the base graph. Then we attach 1 to 3 extra motifs from one of the classes (for example, class A) to the base graph; thus, the number of motifs in class A is more than in class B, representing the graph tends to have the specific property (property A). 

% The ground-truths of a graph are the elements in those motifs including edges (undirected) and self-loop connections. The number of ground-truth elements that are covered by explanation edges is denoted as the hit number. The accuracy is defined as (hit number) / (total number of ground-truth elements).

We also show the time cost of different methods in Table~\ref{tab:time}. Specifically, we report the averaged computation time for 40 graphs in the ClinTox~\cite{moleculenet} dataset.
It shows that our FlowX performs the best with a reasonable time cost. Note that this time cost includes the 500-epoch training.

\subsection{Visualization of Explanation Results}

\begin{figure*}[t]
	\centering
	\includegraphics[width=1\linewidth]{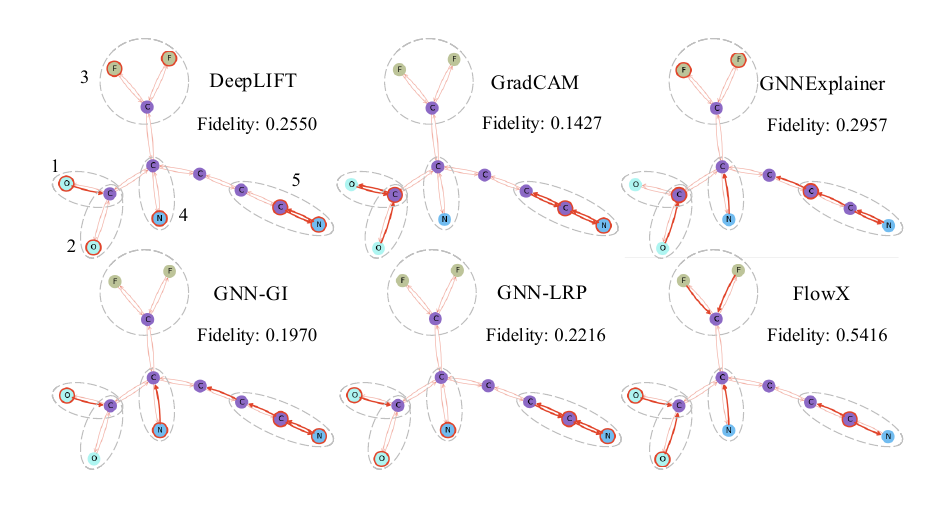}
	 \vspace{-1.2cm}
	\caption{Sample explanation results of different methods.
	Note that explanations are shown as
	directed edges (highlighted red arrow lines) and self-loops (red circles around atoms). In addition, motifs are emphasized by dashed circles and numbered from 1 to 5. FlowX has the most comprehensive important elements choice, which makes it cover all these motifs even with a high Sparsity. The Fidelity in the figure denotes Fidelity+ which targets finding the necessary and comprehensive explanations.}
	\label{fig:vis}
	% \vspace{-0.1cm}
\end{figure*}

Finally, we report the visualization results of different explanation methods in Fig.~\ref{fig:vis}. Specifically, we show the explanations of a molecule graph from ClinTox dataset. The smiles string of this input graph is C(CC(C(F)F)(C(=O)[O-])[NH3+])C[NH3+] and the model to be explained is a 3-layer GCN model. 
In this input graph, motifs 1 and 2 form a Carboxylate Anion; motifs 4 and 5 both contain an Ammoniumyl while motif 3 corresponds to a CF2 group. 
It is clear that our FlowX finds all motifs in the explanation, which directly explains the reason why FlowX can obtain the highest Fidelity+ score. This phenomenon also justifies why necessary explanations are important to the GNN models, \ie, these necessary explanations try to cover all motifs important for the model to make predictions. 
% In addition,  we can conclude that the model combines these motifs for the predictions since we generally observe that explanations with more motifs chosen tend to have higher Fidelity scores. 
Furthermore, by comparing the explanations between DeepLIFT and GNNExplainer, we can extract one more insight that the interactions among different atoms contribute more to the model predictions. Specifically, although DeepLIFT identifies all 5 motifs, its Fidelity score is much lower than GNNExplainer which only identifies four motifs. This indicates the explanation that focuses on the self-loops of atoms is weaker than the one focusing on the one-hop correlations between atoms.

\end{document}